\documentclass{article}

\PassOptionsToPackage{numbers, compress}{natbib}
\usepackage[final]{neurips_2025}
\usepackage{multirow}
\usepackage{makecell}
\usepackage{amsmath}
\usepackage{amssymb}
\usepackage{bbding}
\usepackage{pifont}
\usepackage{wasysym}
\usepackage{tabularx}
\usepackage{utfsym}
\usepackage{fontawesome}
\usepackage{times}
\usepackage{latexsym}

\usepackage{enumitem}
\usepackage{pifont}
\usepackage{fancyvrb}
\newcommand{\cmark}{\ding{51}} 
\newcommand{\xmark}{\ding{55}} 
\usepackage[T1]{fontenc}

\usepackage[utf8]{inputenc}
\usepackage{microtype}

\usepackage[mathscr]{eucal}

\usepackage{inconsolata}
\usepackage{times}
\usepackage{booktabs}    
\usepackage{siunitx}     

\usepackage[T1]{fontenc}
\usepackage[utf8]{inputenc}
\usepackage{mathrsfs}\usepackage{microtype}
\usepackage{inconsolata}
\usepackage{graphicx}
 \usepackage[mathscr]{eucal} 

\usepackage{colortbl}
\usepackage{pgf} 

\definecolor{mygreen}{RGB}{70, 206, 30}
\newcommand{\cellgreen}[1]{%
    \pgfmathsetmacro{\percent}{(#1 - 1) / (35 - 1) * 70 + 15}%
    \edef\temp{\noexpand\cellcolor{mygreen!\percent}\relax #1}%
    \temp
}

\usepackage{hyperref}
\usepackage[capitalize]{cleveref}
\usepackage[most]{tcolorbox}
\tcbset{
  promptstyle/.style={
    colback=gray!5,
    colframe=black!30,
    fonttitle=\bfseries,
    boxrule=0.4pt,
    sharp corners,
    width=\linewidth,
    enhanced,
    breakable,
    before skip=10pt,
    after skip=10pt
  }
}

\definecolor{customcolor_blue}{RGB}{214,228,235}

\usepackage[utf8]{inputenc} 
\usepackage[T1]{fontenc}    
\usepackage{amsfonts}       
\usepackage{nicefrac}       
\usepackage{microtype}      

\usepackage{wrapfig} 

\title{MomentSeeker: A Task-Oriented Benchmark For Long-Video Moment Retrieval}

\begin{document}

\newcommand*\samethanks[1][\value{footnote}]{\footnotemark[#1]}
\maketitle

\setcounter{footnote}{0}

\begin{abstract}

Accurately locating key moments within long videos is crucial for solving long video understanding (LVU) tasks. However, existing benchmarks are either severely limited in terms of video length and task diversity, or they focus solely on the end-to-end LVU performance, making them inappropriate for 
evaluating whether key moments can be accurately accessed. To address this challenge, we propose \textbf{MomentSeeker}, a novel benchmark for long-video moment retrieval (LVMR), distinguished by the following features. First, it is created based on long and diverse videos, averaging over 1,200 seconds in duration, and collected from various domains, e.g., movie, anomaly, egocentric, and sports. Second, it covers a variety of real-world scenarios in three levels: global-level, event-level, and object-level, covering common tasks like action recognition, object localization,  causal reasoning, etc. Third, it incorporates rich forms of queries, including text-only queries, image-conditioned queries, and video-conditioned queries. On top of MomentSeeker, we conduct comprehensive experiments for both generation-based approaches (directly using MLLMs) and retrieval-based approaches (leveraging video retrievers). Our results reveal the significant challenges in long-video moment retrieval in terms of accuracy and efficiency, despite improvements from the latest long-video MLLMs and task-specific fine-tuning. We have publicly released MomentSeeker\footnote{Code is available on this repository: \url{https://yhy-2000.github.io/MomentSeeker/}} to facilitate future research in this area. 
\end{abstract}

\section{Introduction}~\label{sec:intro} 
Long-video understanding (LVU) attracts tremendous attention from the community given its importance to a variety of real-world applications, such as video analysis, embodied AI, and autonomous driving. In recent years, growing efforts have been made to advance the progress of LVU, especially those from multi-modal large language models (MLLMs). Nevertheless, it remains an unsolved problem, primarily due to challenges in both response quality and running efficiency. Long-video understanding requires a diverse set of skills from MLLMs, among which the ability to accurately identify and access key moments is crucial~\cite{wu2024longvideobench}. Although long videos contain rich contextual information, many tasks require reasoning over only a few critical moments. Solving these tasks effectively depends on the accurate retrieval of such moments. In this context, long-video moment retrieval (LVMR) serves a dual purpose: it not only reflects MLLMs' performance on long-video understanding, but also facilitates the development of retrieval methods to address LVU tasks.

However, the advancement of LVMR is severely constrained due to the lack of appropriate evaluation benchmarks. Existing moment retrieval datasets~\cite{qvhighlights, gao2017charadeSTA, lei2020tvr} are insufficient for LVMR, as they are created using short videos, typically only tens of seconds in duration. Besides, many of these datasets use video captions as search queries, which are overly simplistic and differ significantly from real-world LVU tasks. The recent specialized LVU benchmarks, such as Video-MME~\cite{fu2024videomme}, MLVU~\cite{zhou2025mlvu}, and LongVideoBench~\cite{wu2024longvideobench}, are not well-suited for evaluating LVMR as well. Most of these benchmarks focus on the ultimate generation quality, ignoring the assessment of whether key moments are accurately retrieved. While there have been tasks like visual needle-in-a-haystack (V-NIAH)~\cite{zhang2024long}, they rely on synthetic test cases that emphasize frame-level reasoning, which differs fundamentally from the objective of identifying key moments for real-world LVU tasks.

To address the above limitations, we propose \textbf{MomentSeeker}, a novel benchmark for \textit{moment retrieval in long videos}. MomentSeeker enables \textit{task-oriented LVMR evaluation}: given a generic LVU task, e.g., causal reasoning, it assesses whether the key moments relevant to the task can be accurately identified within a long video. MomentSeeker introduces the following key features:

\begin{figure*}[t!]
  \centering
  \includegraphics[width=\linewidth]{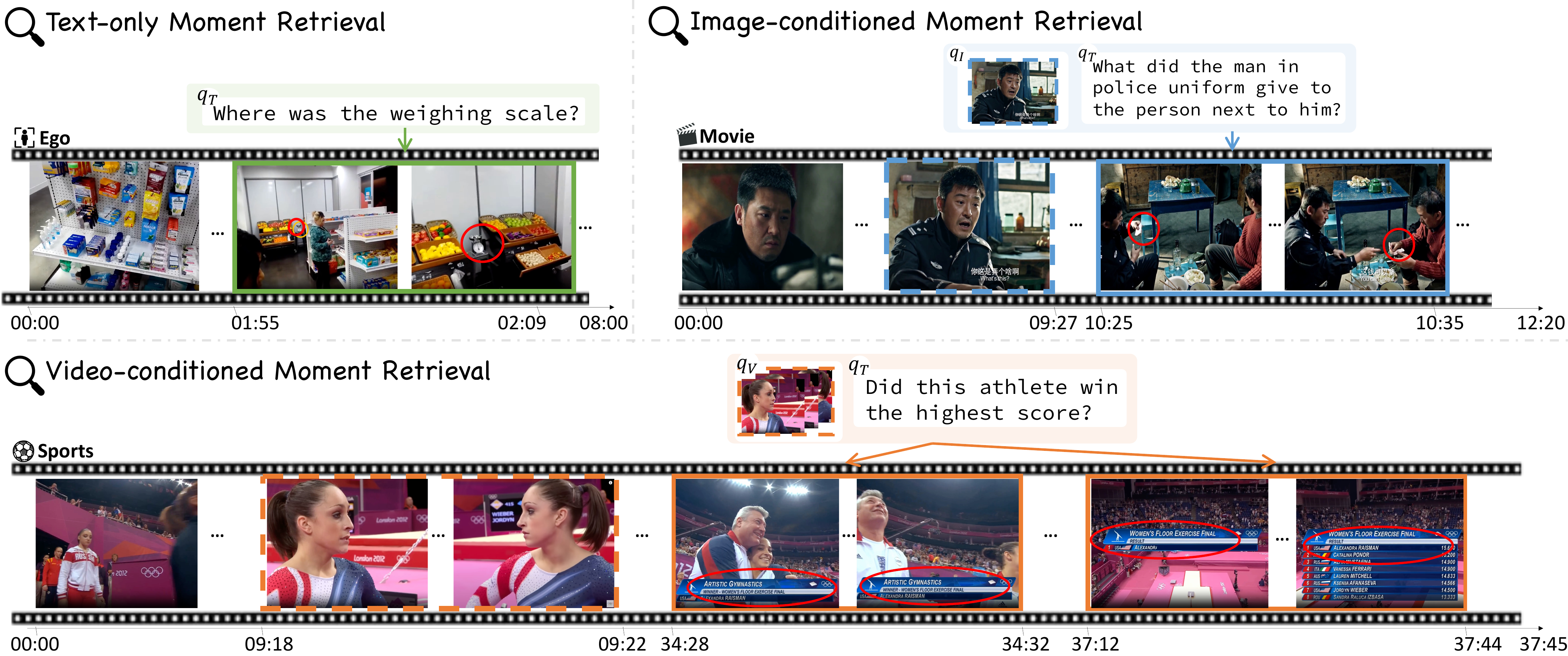}
\caption{Demonstrative examples of the MomentSeeker benchmark. Dashed boxes denote the sources of image $q_I$ and video $q_V$ in the multi-modal queries, while solid boxes indicate the ground truth moment(s). Red circles mark key queried information. }
  \label{fig:intra_demo}
\end{figure*}


$\bullet$ \textbf{Inclusion of long and diverse videos}. MomentSeeker employs long videos while creating its evaluation tasks, with an average duration of more than 1200 seconds. This is significantly longer than those used by existing moment retrieval datasets and comparable to popular LVU benchmarks, such as VideoMME and MLVU. Additionally, the videos are sourced from diverse domains, including movie, sports, anomaly, and egocentric, ensuring broad coverage of real-world scenarios. 


$\bullet$ \textbf{Rich forms of tasks}. MomentSeeker incorporates LVU tasks from three typical scenarios: global-level, event-level, and object-level, which comprehensively cover 9 types of popular tasks, like causal reasoning, spatial reasoning, attribute recognition, OCR, anomaly detection, etc. This sets it apart from existing benchmarks which typically rely on video captions for moment retrieval. Furthermore, MomentSeeker supports three query modalities: text-only queries, image-conditioned queries (i.e., a textual query accompanied by a reference image), and video-conditioned queries (i.e., a textual query with a reference video). These diverse query forms increase the challenge of evaluation and better reflect the complexity of real-world applications.


$\bullet$ \textbf{Fine-grained and high-quality annotation}. For each evaluation task, key moments are annotated with fine-grained timestamps to ensure precision and completeness. Instead of relying on MLLMs or video metadata, which can often be noisy or unreliable, MomentSeeker employs well-informed human annotators to create the tasks, thus ensuring the quality of annotations.  


We conduct comprehensive experiments on MomentSeeker, where two groups of methods are studied. The first employs traditional \textit{retrieval-based methods}~\cite{wei2025uniir,zhu2023languagebind}, which identify key moments by computing the relevance between query and each video chunk. The second comprises \textit{generation-based approaches} which leverage MLLMs to directly predict key moments in long videos. We incorporate diverse models in this group, including both powerful proprietary MLLMs (e.g., GPT-4o) and lightweight open-source alternatives (e.g., Qwen-VL). Our evaluation provides a comprehensive view of existing methods' performance across varying video lengths, domains, and task types, revealing that LVMR remains a challenging problem in both accuracy and efficiency. 

In summary, our contributions are threefold:

1. We introduce MomentSeeker, the first specialized benchmark for long-video moment retrieval. This new benchmark fosters the development of retrieval-based methods and provides a complementary perspective to analyze MLLMs' LVU performance. 

2. MomentSeeker is constructed using long and diverse videos and covers a wide range of LVU tasks. All tasks are annotated with fine-grained, high-quality labels by well-informed human workers. 

3. We perform comprehensive experiments based on MomentSeeker, whose result demonstrates the challenge of LVMR in terms of both accuracy and efficiency.

\section{Related Work}~\label{sec:related_work}

\subsection{Long Video Understanding Benchmarks}


Long video understanding has attracted growing attention, with benchmarks such as VideoMME \cite{fu2024videomme} and MLVU \cite{zhou2025mlvu} supporting tasks on videos ranging from 3 to 60 minutes. LongVideoBench \cite{wu2024longvideobench} highlights the need for fine-grained multi-modal retrieval and reasoning by introducing referring reasoning QA tasks. However, these benchmarks evaluate long video understanding capabilities in an end-to-end manner while overlooking in-depth evaluation of moment retrieval. 
To assess MLLM's ability to locate key information within long contexts, V-NIAH~\cite{zhang2024long} proposes a needle-in-a-haystack benchmark by inserting synthetic frames into long videos. However, due to its synthetic nature, it fails to reflect real-world multi-modal complexity and temporal dependencies. LV-Haystack \cite{ye2025lvhaystack} is a concurrent benchmark that focuses on locating task-aware keyframes in long videos. Nevertheless, it emphasizes frame-level matching and is limited in both task variety and scenario diversity.

While MomentSeeker adopts a task taxonomy resembling existing long video understanding tasks, such as Causal Reasoning, this design aims to maintain interpretability and continuity with prior work rather than mere reformulation. In contrast to previous long video understanding benchmarks that only assess answer correctness, MomentSeeker explicitly verifies grounding by requiring models to retrieve the exact temporal moment supporting each query. This addresses a key limitation: correct answers do not necessarily imply correct grounding, as models may guess without truly leveraging visual evidence. Empirically, even strong models that perform well on long video understanding benchmarks, such as Qwen2.5-VL-72B achieving 79.1 on VideoMME, obtain only 17.2 R@1 on MomentSeeker, revealing the difficulty of true moment localization. By demanding timestamp prediction, MomentSeeker enhances interpretability and factuality, making model reasoning transparent and verifiable. Moreover, accurate moment retrieval facilitates downstream long-video reasoning and agentic tasks. Overall, MomentSeeker targets the overlooked challenge of long-video moment retrieval, requiring deeper understanding, precise localization, and transparent reasoning, thus advancing the long video understanding field.


\subsection{Moment Retrieval Benchmarks}

Moment retrieval has been extensively studied in short-video scenarios~\cite{qvhighlights,gao2017charadeSTA,THUMOS14,hendricks2017didemo,lei2020tvr}, where videos typically last under three minutes and require limited temporal reasoning. Most existing benchmarks rely on synthetic settings (e.g., subtitle-based retrieval) or domain-specific content (e.g., TV shows or human activities), which restricts their applicability to real-world situations.
Recent multimodal query grounding studies~\cite{ref1,ref2} extend this line of work but still focus on short clips and rely primarily on template-based or text-only queries. Specifically,~\cite{ref1} employs automatically generated textual queries, while~\cite{ref2} incorporates audio, subtitles, and knowledge-base information to improve multimodal understanding, yet it does not address fine-grained temporal grounding. In contrast, our \textbf{MomentSeeker} benchmark targets long-video moment retrieval, featuring expert-authored, high-quality queries that require deeper temporal and causal reasoning.
Additionally, long-form video benchmarks such as Ego4D and Ego-Exo4D~\cite{grauman2022ego4d,grauman2024egoexo4dunderstandingskilledhuman} contribute valuable egocentric datasets but provide only clip-level captions centered on local physical actions, lacking broader scene diversity. MomentSeeker instead spans diverse domains (e.g., sports, movies, surveillance) and introduces multimodal queries (text, text+image, text+video), enabling a more comprehensive and challenging evaluation of long-video understanding.

\section{MomentSeeker}~\label{sec:momentseeker}
\subsection{Task Definition}
The long-video moment retrieval (LVMR) task is formally defined as follows. Given the query for a LVU task $\textbf{q}$, the model is required to predict the key moments within the long video, denoted as $\mathcal{P} = [p^{(1)}, ..., p^{(k)}]$, where $p^i$ stands for the starting and terminating timestamp of the $i$-th moment, i.e., $p^i_s$ and $p^i_t$. The prediction result is measured by its consistency with the ground-truth moments $\mathcal{G} = [g^{(1)}, ..., g^{(m)}]$: $\phi(\{sim(p^i, g^i)|i=1,...,m\})$, where $sim(\cdot)$ and $\phi(\cdot)$ are the predefined similarity function and integration function, respectively (to be introduced in Section \ref{sec:eval_metrics}).

\subsection{Video Collection}
To more comprehensively evaluate the LVMR task, we construct a video dataset that is both diverse in scenarios and sufficiently long in duration, aiming to better reflect real-world conditions.
To maximize diversity, we curate videos from a wide array of domains, including real-world recordings such as egocentric videos and sports, cinematic productions like movies, and simulated content like cartoons and surveillance videos for anomaly detection. Sports videos, movies, egocentric videos, and cartoons are all high-resolution, each exceeding 1080p. The sports videos are primarily sourced from Olympic tournaments. Although surveillance videos are typically low in resolution (around 320×240) due to limitations of available sources on the Internet, we apply strict filtering to retain only clips with clearly visible abnormal events, ensuring both quality and relevance. Compared to prior benchmarks that are confined to specific domains, such as TV shows~\cite{lei2020tvr}, sports activities~\cite{gao2017charadeSTA,THUMOS14},  our dataset covers a broader range of visual scenarios, reflecting a wider spectrum of content, including real-world, cinematic, and simulated environments.

Furthermore, our benchmark covers a broad range of durations~(Figure~\ref{fig:dataset_statistics} (b)), with the longest videos extending up to approximately one hour. As shown in Table~\ref{tab:dataset_comparison}, our benchmark boasts an average video length of 1201.9 seconds, significantly outpacing previous moment retrieval benchmarks and aligning with the LVU benchmark. The meticulous video collection process guarantees both domain diversity and comprehensive coverage of video durations, strengthening the benchmark's ability to support LVMR tasks for the first time and its utility for real-world applications.

\subsection{Task Creation}

\begin{figure*}[t!]
  \centering

  \includegraphics[width=\linewidth]{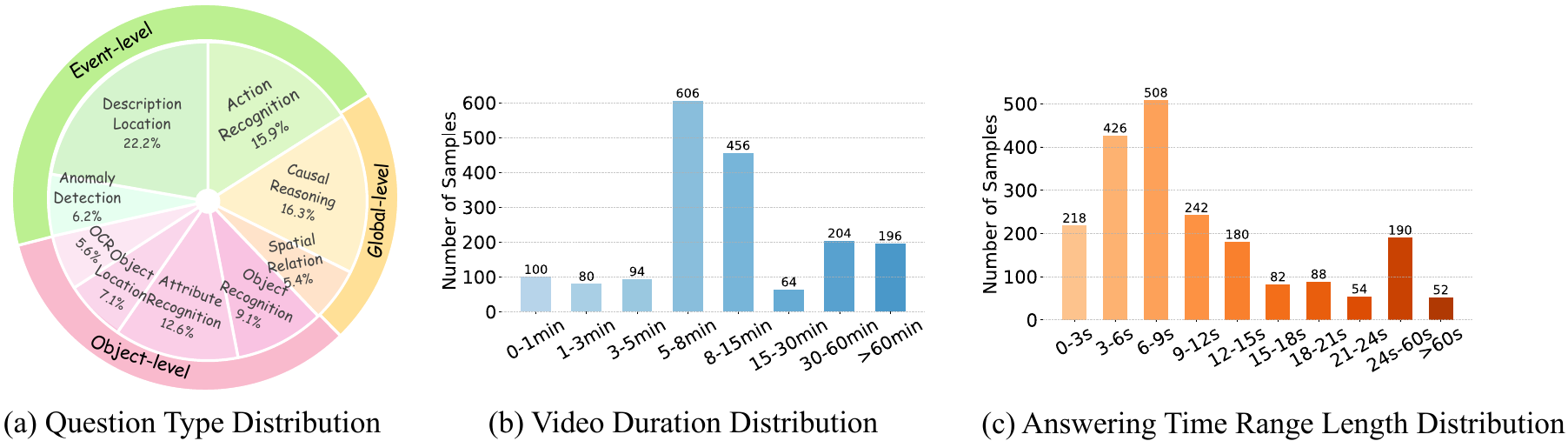}
\caption{Dataset statistics. (a). Question type distribution, (b). Video duration distribution across samples, and (c) Answering time range length distribution across samples. MomentSeeker has a full spectrum of video length and covers different core abilities of the moment retrieval task.}
  \label{fig:dataset_statistics}
\end{figure*}

To evaluate video moment retrieval at different levels of semantic granularity, we construct the MomentSeeker benchmark with a hierarchical task taxonomy, as illustrated in Figure~\ref{fig:dataset_statistics} (a). The benchmark encompasses a wide range of question types, each targeting distinct perception or reasoning capabilities. We categorize all tasks into three levels: global, event, and object, based on the semantic scope of the queried information. While all tasks share the common objective of grounding queries to precise temporal intervals in videos, they differ significantly in the level of reasoning, contextual integration, and perceptual detail required.

\paragraph{Global-Level Moment Retrieval.}
Global-level tasks assess the model's ability to reason over extended temporal contexts and to understand high-level semantics that span large portions of the video. These tasks often require modeling causal, temporal, or spatial relationships between multiple events or scenes.
\textit{1). Causal Reasoning} requires the model to uncover causal relationships between the question event and the answer event in order to locate the latter. For instance, answering ``Why does the man need to close the bedroom window?'' involves identifying a temporally distant but causally related prior event such as ``It is snowing outside.''
Similarly, \textit{2). Spatial Relation} requires the model to infer spatial relationships between the question event and the answer event. Questions like ``How many people are there opposite the man sitting on the sofa?'' demand holistic scene understanding and spatial reasoning. These tasks require integrating information across scenes, making them the most contextually demanding category.

\begin{table}[t!]
\centering
\resizebox{\textwidth}{!}{
\begin{tabular}{lccccccc}
\toprule
\textbf{Benchmark} & \textbf{Label} & \makecell[c]{\textbf{Moment-}\\\textbf{targeted?}} & \makecell[c]{\textbf{Task-}\\\textbf{oriented?}} & \textbf{Avg. Dur (s)} & \textbf{\#Videos} & \textbf{\#Queries} & \textbf{Domain} \\
\midrule
\multicolumn{8}{l}{\textbf{Moment retrieval benchmarks}} \\[0.1cm]
TVR~\cite{lei2020tvr} & Auto & \cmark & \xmark & 76.2 & 1090 & 5450 & TV show \\
CharadesSTA~\cite{gao2017charadeSTA} & Human & \cmark & \xmark & 30.6 & 1334 & 3720 & Activity \\
THUMOS14~\cite{THUMOS14} & Human & \cmark & \xmark & 186.4 & 216 & 3457 & Action \\
QVHighlights~\cite{qvhighlights} & Human & \cmark & \cmark & 150 & 476 & 1542 & Vlog/News \\
\midrule
\multicolumn{8}{l}{\textbf{LVU benchmarks}} \\[0.1cm]
VideoMME~\cite{fu2024videomme} & Human & \xmark & \cmark & 1021.3 & 900 & 2700 & YouTube \\
MLVU~\cite{zhou2025mlvu} & Human & \xmark & \cmark & 905.8 & 349 & 502 & Open \\
LongVideoBench~\cite{wu2024longvideobench} & Human & \xmark & \cmark & 574.9 & 753 & 1337 & Open \\
V-NIAH~\cite{zhang2024long} & Auto & \xmark & \cmark & - & - & 5 & Open \\
\midrule
\textbf{MomentSeeker} & Human & \cmark & \cmark & 1201.9 & 268 & 1800 & Open \\
\bottomrule
\end{tabular}
}
\vspace{0.1cm}
\caption{Comparison of popular moment retrieval benchmarks, LVU benchmarks (with test set statistics), and our proposed MomentSeeker benchmark. \textit{Avg. Dur} denotes average video duration.}
\label{tab:dataset_comparison}
\end{table}

\begin{figure*}[t!]
\includegraphics[width=\linewidth]{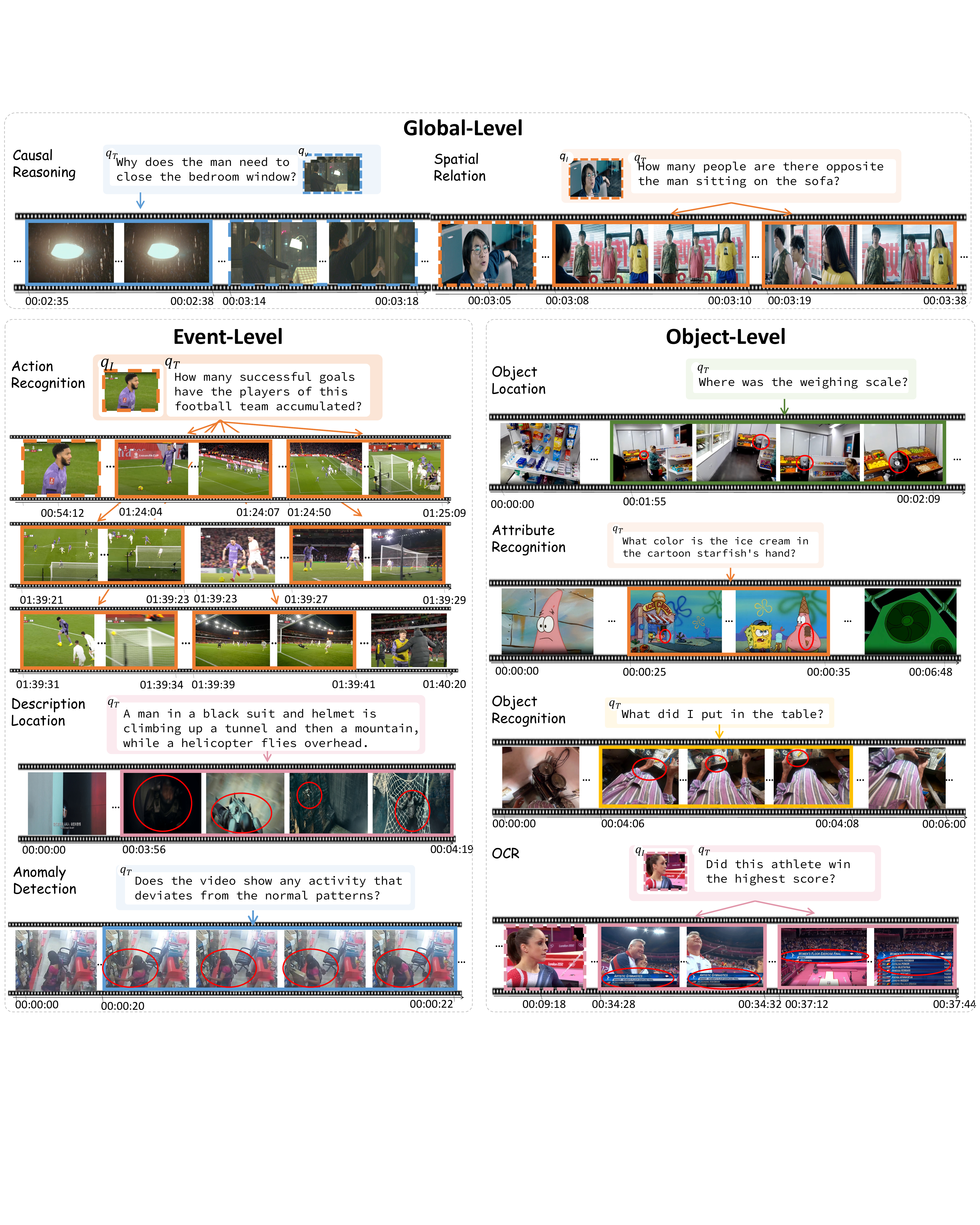}
\caption{Examples of each task. Dashed boxes show sources of query image $q_I$ and video $q_V$; solid boxes mark ground truth moments. Red circles highlight key queried information.}
\label{fig:bench_examples}
\vspace{-0.3cm}
\end{figure*}

\paragraph{Event-Level Moment Retrieval.}
Event-level tasks aim to retrieve moments corresponding to specific actions or events, requiring more localized reasoning than global tasks. The event in the query is exactly what needs to be localized. This category includes:
\textit{1) Description Location}, which matches detailed textual descriptions to video segments, focusing on visual-text alignment with minimal reasoning (e.g., ``A man in a black suit and helmet is climbing up a tunnel...'').
\textit{2) Action Recognition}, which involves identifying and classifying specific actions, such as counting successful goals in a football match.
\textit{3) Anomaly Detection}, which requires detecting abnormal events without explicit textual cues~(e.g., ``Does the video show any activity that deviates from the normal patterns?''). This task relies on the model's ability to infer irregularities.
 Overall, event-level tasks target a specific event and strike a balance between the long-range semantics of global tasks and the fine-grained specificity of object-centric tasks.

\paragraph{Object-Level Moment Retrieval.}
Object-level tasks focus on identifying specific objects, their attributes, or spatio-temporal positions, emphasizing fine-grained perception and low-level visual understanding. Subtypes include:
\textit{1) Object recognition} (e.g., ``What did I put on the table?''), identifying entities in the scene;
\textit{2) Object localization} (e.g., ``Where was the weighing scale?''), grounding objects in space and time;
\textit{3) Attribute classification} (e.g., ``What color is the ice cream in the cartoon starfish's hand?''), recognizing detailed visual properties;
\textit{4) OCR-based reasoning} (e.g., ``Did this athlete win
the highest score?''), detecting, reading, and analyzing the embedded text.
These tasks typically span short durations and demand high spatial accuracy, contrasting with the broader semantic focus of global and event-level tasks.

By organizing tasks in a hierarchical structure, the MomentSeeker benchmark provides a unified yet granular framework for evaluating the diverse capabilities of moment retrieval methods. Additionally, considering that humans often convey information using a combination of modalities (e.g., ``Why does the man in this image need to close the bedroom window?''), such multi-modal expressions tend to be more natural and accurate. Therefore, in this benchmark, we incorporate three modality combinations: text-only, text with image, and text with video, with text-only serving as the primary modality and multi-modal queries acting as auxiliary. Examples of each task are shown in Figure~\ref{fig:bench_examples}.

\subsection{Data Annotation}
We formally define answering moments as the minimal collection of temporally localized segments that constitute both necessary and sufficient visual evidence to answer a given question. For single-detail questions (e.g., ``What color is the ice cream in the cartoon starfish's hand?''), a single concise moment suffices. In contrast, multi-detail questions (e.g., ``How many successful goals have the players of this football team accumulated?'') span multiple, non-redundant moments, which together constitute both necessary and sufficient evidence to answer the question. As for how models identify the segments, we do not impose any constraints. For example, retrieval-based approaches may select from pre-segmented clips, while generative methods may directly predict the temporal boundaries.

Building on this foundation, we design a meticulous annotation protocol. Unlike prior datasets relying on subtitles~\cite{lei2020tvr} or coarse action labels~\cite{THUMOS14,gao2017charadeSTA}, we engage expert annotators to craft context-rich, natural questions for LVMR. Each question is paired with precise answer intervals and may be text-only, image-conditioned, or video-conditioned (examples are shown in Figure~\ref{fig:bench_examples}). In the image-conditioned setting, annotators provide both a textual question and a representative frame. In the video-conditioned setting, temporal query windows serve as inputs. In terms of quantity, we ensure that text-only queries constitute the majority, with image- and video-conditioned queries serving as auxiliary, resulting in a final distribution ratio of 5:2:2. To scale annotation efficiently without compromising quality, we leverage a strong multi-modal model~\cite{qwen2vl} to generate detailed descriptions of video clips for the ``Description Location'' task, which are then curated and refined by human annotators. After that, all samples are assigned one of nine predefined task types, including action recognition, spatial relations, causal reasoning, and others. For quality control, we first apply rule-based filtering to remove noisy samples (e.g., redundant queries, invalid intervals). Then, annotators conduct cross-checking to ensure question clarity and label validity. This two-pass process ensures consistent and reliable annotations. Figure~\ref{fig:dataset_statistics} presents detailed statistics, illustrating the diverse task types, broad video durations, and varied answering times, demonstrating the benchmark's robustness and versatility for complex LVMR tasks.

\subsection{Evaluation Metrics}\label{sec:eval_metrics}
In moment retrieval tasks, the answer to a question may involve either a single moment or multiple moments in the video. To provide a comprehensive evaluation across both single- and multi-detail scenarios, we use Recall@1 and Mean Average Precision@5 as complementary metrics.

\textbf{Recall@1 (R@1)} is a standard metric in moment retrieval~\cite{qvhighlights,hendricks2017didemo,gao2017charadeSTA}, measuring whether the top-ranked prediction $p^{(1)}$ matches any ground-truth moment $g^j \in \mathcal{G}$ with similarity $sim(p^{(1)}, g^j)$ exceeding a predefined threshold. The similarity function $sim$ is typically defined as Intersection over Union (IoU)~\cite{qvhighlights,hendricks2017didemo}, which measures the overlap between the prediction and the ground-truth moment. Accordingly, the set $\{sim(p^{(1)}, g^j)\}$ contains the similarity scores between the prediction and all ground-truth moments. The function $\phi(\cdot)$ returns 1 if at least one of these scores exceeds the threshold, and 0 otherwise. This design accounts for the presence of multiple valid ground-truth moments, treating the prediction as correct if it sufficiently overlaps with any one of them. The final R@1 score is obtained by averaging over all queries.

\textbf{Mean Average Precision@5 (mAP@5)} evaluates both the accuracy and the ranking quality of the top-5 predicted temporal segments for each query. It complements Recall@1, which only considers top-1 accuracy, and is widely used in scenarios involving multiple ground-truth moment retrieval~\cite{qvhighlights}. For each query, predictions are ranked by confidence, and a prediction is considered correct if its temporal IoU with any unmatched ground-truth segment exceeds a threshold. Each ground-truth can be matched at most once. The Average Precision (AP) is computed by averaging the precision at each correct prediction within the top-5 ranked results. The final mAP@5 is obtained by averaging the AP across all queries. mAP@5 reflects not only whether relevant moments are retrieved, but also how well they are ranked, rewarding systems that return correct predictions earlier in the list.

\section{Experiments}\label{sec:experiments} 

\subsection{Experiment Setting \& Baselines}
In this paper, we consider two mainstream approaches to solving the LVMR problem: retrieval-based methods and generation-based methods. Retrieval-based methods rely on chunking, where the entire video is divided into multiple independent segments. By computing the similarity between each segment and the question, these methods rank all candidate intervals and identify the most relevant ones. Such methods are often used as the retrieval component in RAG-based LVU models~\cite{yuan2025memory,ataallah2024goldfish,luo2024videorag}, where a small set of relevant video clips is first retrieved based on the question and then passed to downstream MLLMs to complete the LVU task. The second category is generation-based methods, which aim to directly test the temporal reasoning ability of MLLMs. These methods feed both the question and the video into the MLLM, prompting it to directly output a list of time intervals containing the answer in an end-to-end fashion. This evaluation of temporal grounding capability has been highlighted in several state-of-the-art MLLMs~\cite{bai2025qwen25vl,guo2025seed15vltechnicalreport}. In the main experiment, we set the IoU threshold to 0.3. Additionally, results with different IoU thresholds are provided in Appendix~\ref{app:iou}. Below, we detail the settings and baseline models for each category.

\paragraph{Retrieval-based Methods.}
We divide each video into fixed 10-second chunks without tailoring the segmentation to any specific model. The IoU threshold is set to 0.3 in the main results, with additional thresholds reported in the appendix. We compare mainstream retrieval-based methods, including dual-encoder models such as InternVideo2~\cite{wang2025internvideo2} and LanguageBind~\cite{zhu2023languagebind}. Compositional retrieval methods such as COVR~\cite{ventura2024covr} and MM-RET~\cite{zhou2024megapairs} generate multi-modal embeddings by combining vision and text inputs. Recent approaches like E5V~\cite{jiang2024e5} and VLM2VEC~\cite{jiang2025vlm2vec} utilize MLLMs to embed arbitrary combinations of modalities. All these models first generate query embeddings and embeddings for pre-segmented video clips, compute similarity scores between them, rank the video clips based on these scores, and return the top-$k$ clips as the predicted moments. Details on how each model generates embeddings are provided in the Appendix~\ref{app:retrieval_setting}.

\begin{table*}[t!]
\centering
\caption{Main results across different meta-tasks. \#Frames indicates the number of input frames for generation-based methods and per-clip frames for retrieval-based methods.}
\label{tab:main_exp}
\resizebox{\textwidth}{!}{%
\begin{tabular}{
  l|cc|
  cc@{\hskip 2pt}>{\columncolor{white}}c@{\hskip 2pt}  
  cc@{\hskip 2pt}>{\columncolor{white}}c@{\hskip 2pt}
  cc@{\hskip 2pt}>{\columncolor{white}}c@{\hskip 2pt}
  cc}
\toprule
\multirow{2}{*}{\textbf{Method}} & 
\multirow{2}{*}{\textbf{\#Size}} & 
\multirow{2}{*}{\textbf{\#Frames}} & 
\multicolumn{2}{c}{\textbf{Global-level}} & &
\multicolumn{2}{c}{\textbf{Event-level}} & &
\multicolumn{2}{c}{\textbf{Object-level}} & &
\multicolumn{2}{c}{\textbf{Overall}} \\

\cmidrule{4-5} \cmidrule{7-8} \cmidrule{10-11} \cmidrule{13-14}  
 & & & R@1 & mAP@5 & 
\multicolumn{1}{c}{} & 
R@1 & mAP@5 & 
\multicolumn{1}{c}{} & 
R@1 & mAP@5 & 
\multicolumn{1}{c}{} & 
R@1 & mAP@5 \\

\midrule
\multicolumn{14}{c}{\textit{\textbf{Generation-based Methods}}} \\
\hline
GPT-4o(2024-11-20)~\cite{openai2023gpt4} & - & 128  & \cellgreen{12.7} & \cellgreen{12.7} & & \cellgreen{21.3} & \cellgreen{22.2} & & \cellgreen{20.4} & \cellgreen{21.5} & & \cellgreen{18.2} & \cellgreen{18.9} \\
Gemini-2.5-Pro~\cite{Kavukcuoglu2025gemini2.5} & - & 128 & \cellgreen{20.5} & \cellgreen{22.5}&  &\cellgreen{31.7}& \cellgreen{33.9}& & \cellgreen{35.2}& \cellgreen{36.3}& & \cellgreen{29.6}& \cellgreen{31.4} \\
TimeChat~\cite{ren2024timechat}& 7B & 96 & \cellgreen{2.6} & \cellgreen{2.6} & & \cellgreen{6.7} & \cellgreen{6.7} & & \cellgreen{4.4} & \cellgreen{4.4} & & \cellgreen{5.9} & \cellgreen{5.9} \\
Lita~\cite{huang2024lita} & 13B & 100 & \cellgreen{2.6} & \cellgreen{2.6} & & \cellgreen{7.2} & \cellgreen{7.2} & & \cellgreen{1.8} & \cellgreen{1.8} & & \cellgreen{5.6} & \cellgreen{5.6} \\
Qwen2.5VL~\cite{bai2025qwen25vl} & 7B & 768 & \cellgreen{4.6} & \cellgreen{3.8} & & \cellgreen{12.0} & \cellgreen{12.2} & & \cellgreen{4.3} & \cellgreen{4.2} & & \cellgreen{8.1} & \cellgreen{8.0} \\
InternVL3~\cite{zhu2025internvl3} & 8B & 96 & \cellgreen{3.9} & \cellgreen{3.5} & & \cellgreen{7.8} & \cellgreen{8.5} & & \cellgreen{4.1} & \cellgreen{4.1} & & \cellgreen{5.9} & \cellgreen{6.1} \\
Eagle2.5~\cite{chen2025eagle25boostinglongcontext} & 8B & 256 & \cellgreen{9.3} & \cellgreen{9.2} && \cellgreen{9.3} & \cellgreen{9.4} && \cellgreen{7.2} & \cellgreen{7.4} && \cellgreen{8.7} & \cellgreen{8.7} \\
VideoChat-Flash~\cite{li2025videochatflash} & 7B & 256 & \cellgreen{2.9} & \cellgreen{3.1} && \cellgreen{9.4} & \cellgreen{9.4} && \cellgreen{7.2} & \cellgreen{7.2} && \cellgreen{7.3} & \cellgreen{7.4} \\
Video-LLaMA3~\cite{zhang2025videollama3frontiermultimodal} & 7B & 256 & \cellgreen{11.1} & \cellgreen{9.9} && \cellgreen{20.9} & \cellgreen{19.0} && \cellgreen{12.8} & \cellgreen{11.7} && \cellgreen{16.4} & \cellgreen{14.9} \\
InternVL3~\cite{zhu2025internvl3} & 38B & 96 &  \cellgreen{11.1} & \cellgreen{10.5} & & \cellgreen{20.8} & \cellgreen{21.2} & & \cellgreen{11.3} & \cellgreen{11.5} & & \cellgreen{15.8} & \cellgreen{16.0} \\
LLaVA-Video~\cite{zhang2024llavavideo} & 72B & 96  & \cellgreen{3.6} & \cellgreen{3.5} & & \cellgreen{8.6} & \cellgreen{9.8} & & \cellgreen{4.6} & \cellgreen{5.6} & & \cellgreen{6.3} & \cellgreen{7.2} \\
Qwen2.5VL~\cite{bai2025qwen25vl}& 72B & 768 & \cellgreen{13.6} & \cellgreen{13.0} & & \cellgreen{21.9} & \cellgreen{21.8} & & \cellgreen{12.2} & \cellgreen{11.9} & & \cellgreen{17.2} & \cellgreen{16.9} \\
\hline
\multicolumn{14}{c}{\textit{\textbf{Retrieval-based Methods}}} \\
\hline
E5V~\cite{jiang2024e5} & 8.4B & 1 & \cellgreen{13.1} & \cellgreen{19.5} & & \cellgreen{14.5} & \cellgreen{20.7} & & \cellgreen{14.9} & \cellgreen{19.8} & & \cellgreen{14.3} & \cellgreen{20.1} \\
UniIR~\cite{wei2025uniir} & 428M & 1  & \cellgreen{14.9} & \cellgreen{19.4} & & \cellgreen{11.5} & \cellgreen{17.9} & & \cellgreen{8.2} & \cellgreen{13.9} & & \cellgreen{11.2} & \cellgreen{16.9} \\
MM-Ret~\cite{zhou2024megapairs} & 148M & 1 & \cellgreen{14.2} & \cellgreen{17.9} & & \cellgreen{13.6} & \cellgreen{19.4} & & \cellgreen{9.7} & \cellgreen{15.4} & & \cellgreen{12.4} & \cellgreen{17.7} \\
CoVR~\cite{ventura2024covr} & 588M & 15 & \cellgreen{9.8} & \cellgreen{15.4} & & \cellgreen{13.7} & \cellgreen{19.9} & & \cellgreen{14.4} & \cellgreen{18.9} & & \cellgreen{13.0} & \cellgreen{18.5} \\
LanguageBind~\cite{zhu2023languagebind} & 428M & 8 & \cellgreen{16.2} & \cellgreen{24.6} & & \cellgreen{21.4} & \cellgreen{29.4} & & \cellgreen{15.5} & \cellgreen{21.0} & & \cellgreen{18.2} & \cellgreen{25.4} \\
InternVideo2~\cite{wang2025internvideo2} & 1B & 8 & \cellgreen{16.8} & \cellgreen{24.5} & & \cellgreen{23.5} & \cellgreen{30.9} & & \cellgreen{17.0} & \cellgreen{22.7} & & \cellgreen{19.7} & \cellgreen{26.6} \\
\bottomrule
\end{tabular}
}
\end{table*}

\begin{figure*}[t!]
    \centering
    \includegraphics[width=\linewidth]{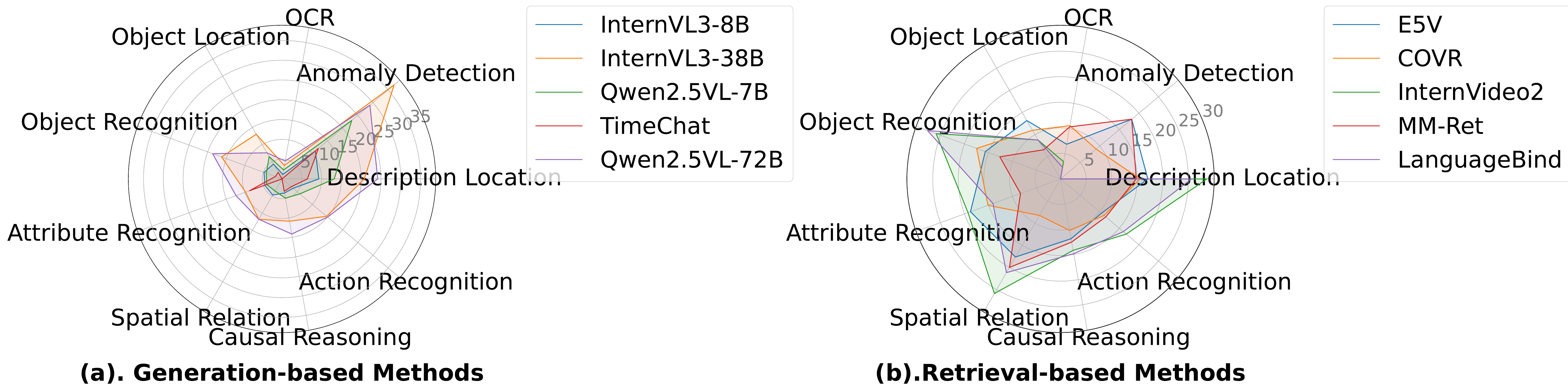}
  \caption{Sub-task performance of different retrieval-based methods and generation-based methods.}
    \label{fig:task_exp}
\end{figure*}

\paragraph{Generation-based Methods.}
For the experimental setup, we input the video (either uniformly down-sampled frames or raw mp4) into the model and provide it with the total video duration as well as the timestamp corresponding to each frame. We then prompt the model to output a list of one or more time intervals containing the answer. The number of frames follows the optimal settings officially recommended by each model, and the time instruction format also follows the official input specifications (details for each model can be found in the Appendix~\ref{app:generation_setting}). Additionally, we provide the ablation study of different frame numbers in the Appendix~\ref{app:frame_result}. We adopt the same IoU-based evaluation as used for the retrieval-based methods to enable a fair comparison. For baselines, we evaluate a wide range of state-of-the-art MLLMs, including models specifically designed for moment retrieval (e.g., TimeChat~\cite{ren2024timechat} and Lita~\cite{huang2024lita}), leading open-source multi-modal models such as LLaVA-Video~\cite{zhang2024llavavideo}, InternVL3~\cite{zhu2025internvl3}, and Qwen2.5VL~\cite{bai2025qwen25vl}, as well as long-video-oriented MLLMs including VideoLLaMA3~\cite{zhang2025videollama3frontiermultimodal}, Eagle2.5~\cite{chen2025eagle25boostinglongcontext} and VideoChat-Flash~\cite{li2025videochatflash}. Additionally, we evaluate powerful closed-source models GPT-4o~(2024-11-20)~\cite{openai2023gpt4} and Gemini-2.5-Pro~\cite{Kavukcuoglu2025gemini2.5} to provide a comprehensive assessment.

\subsection{Main Results}
Table~\ref{tab:main_exp} and Figure~\ref{fig:task_exp} shows the overall results. Results indicate that despite advances in video understanding,  \textbf{LVMR remains challenging for current models.}  For example, InternVL3-8B achieves only 5.9 R@1 and 6.1 mAP@5, while even large-scale models like Qwen2.5VL-72B (handling up to 768 frames) reach just 17.2 R@1, underscoring current MLLMs' limitations in fine-grained temporal perception. Retrieval-based methods perform slightly better but still fall short. For example, the MLLM-based retriever E5V reaches 14.3 R@1. Even the state-of-the-art retriever InternVideo2 achieves only 19.7 R@1 in different settings. This subpar performance reflects that most retrievers are tuned for direct cross-modal alignment (e.g., caption-based retrieval), while our task requires deeper instruction understanding and complex multi-modal reasoning. Some representative visualization cases are provided in Appendix~\ref{app:visual}.

\subsection{Analysis}

\begin{figure*}[t!]
\centering
\begin{minipage}{0.5\linewidth}
    \centering
    \includegraphics[width=\linewidth]{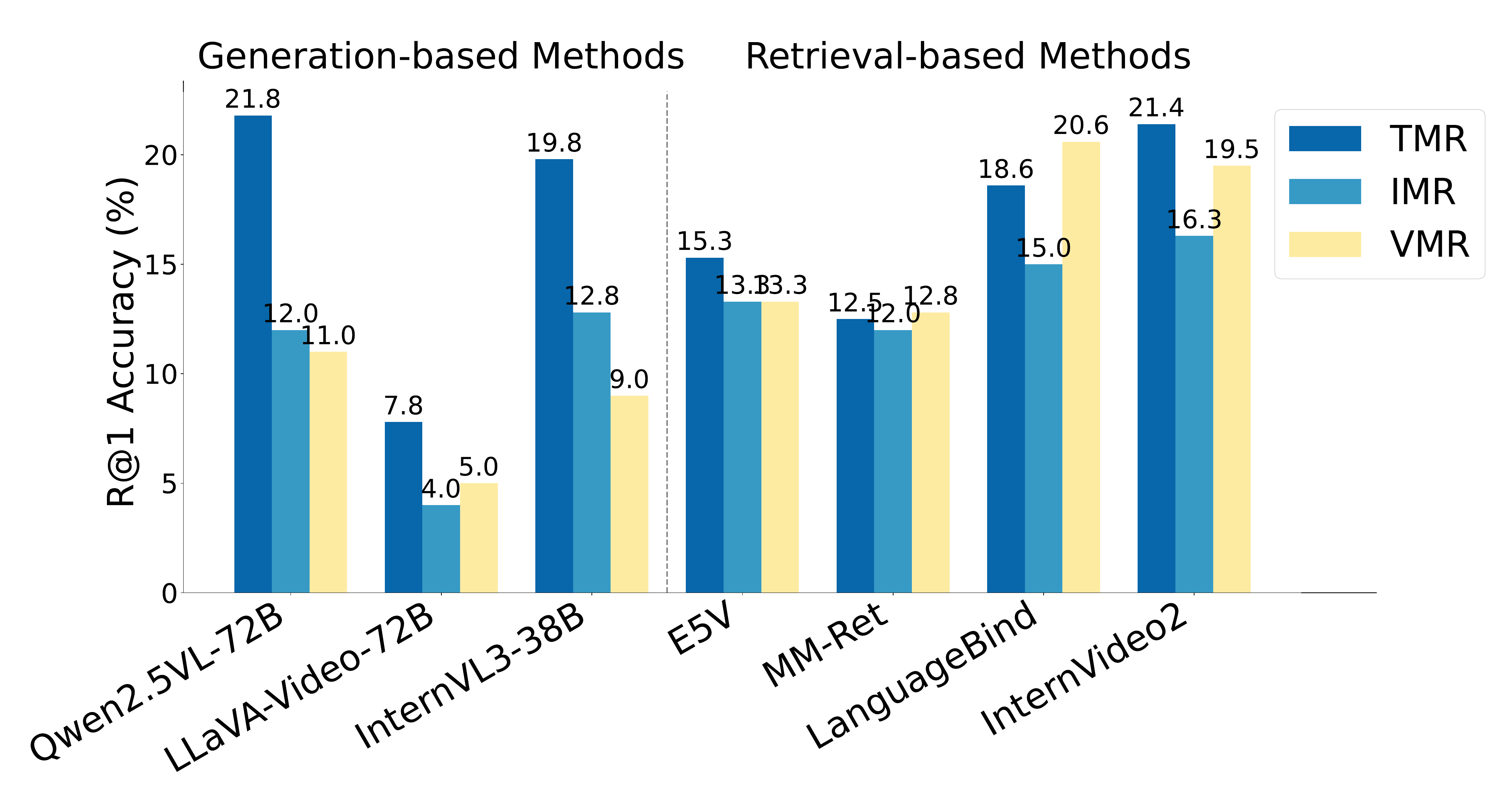}
\caption{Evaluation results w.r.t. query modalities: TMR (text-only), IMR (image-conditioned), and VMR (video-conditioned) moment retrieval.}
    \label{fig:modality}
\end{minipage}
\hfill
\begin{minipage}{0.45\linewidth}
    \centering
    \includegraphics[width=\linewidth]{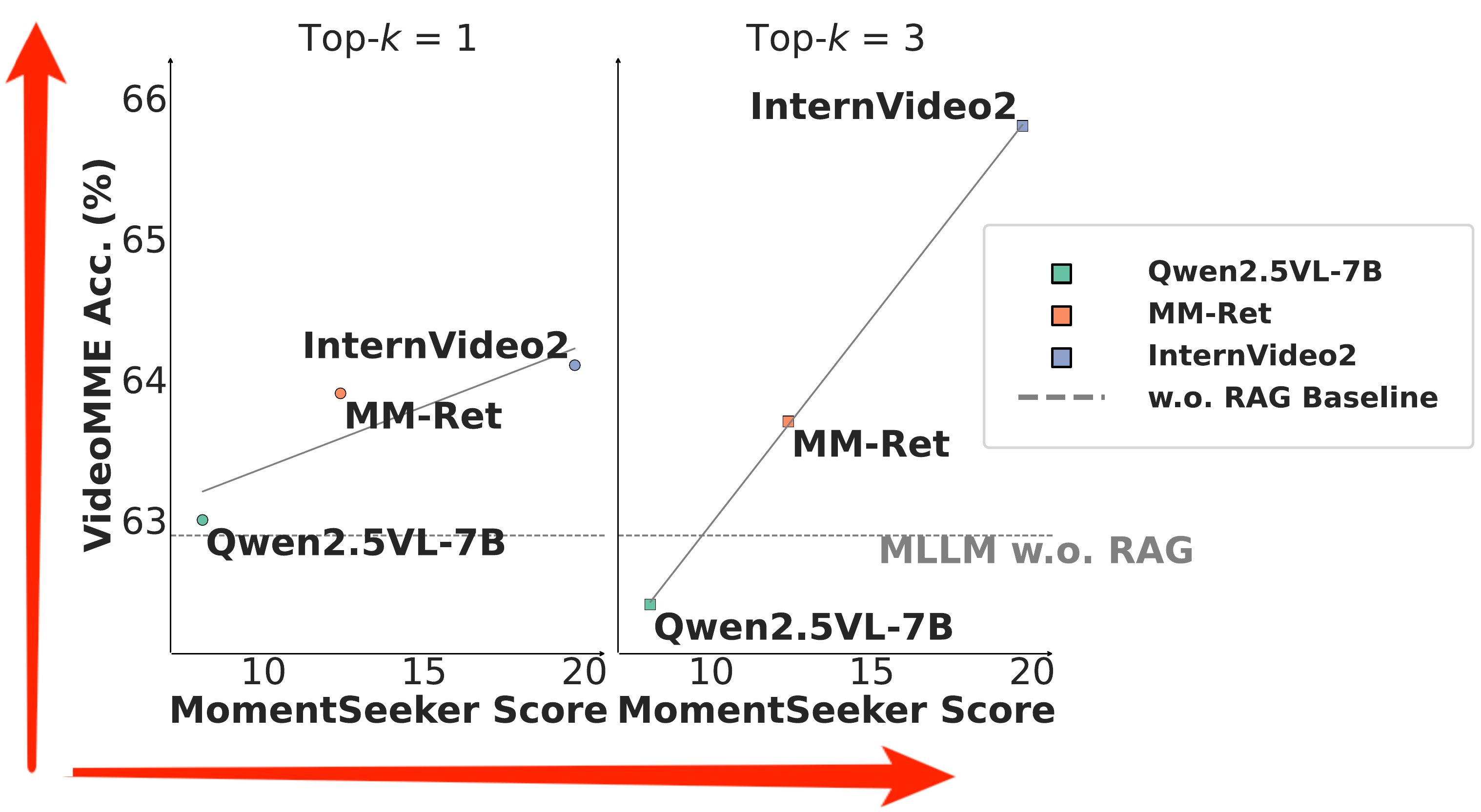}
    \caption{Comparison of MomentSeeker and LVU performance. }
    \label{fig:retriever}
\end{minipage}
\end{figure*}

\begin{figure*}[t!]
  \centering
  \includegraphics[width=\linewidth]{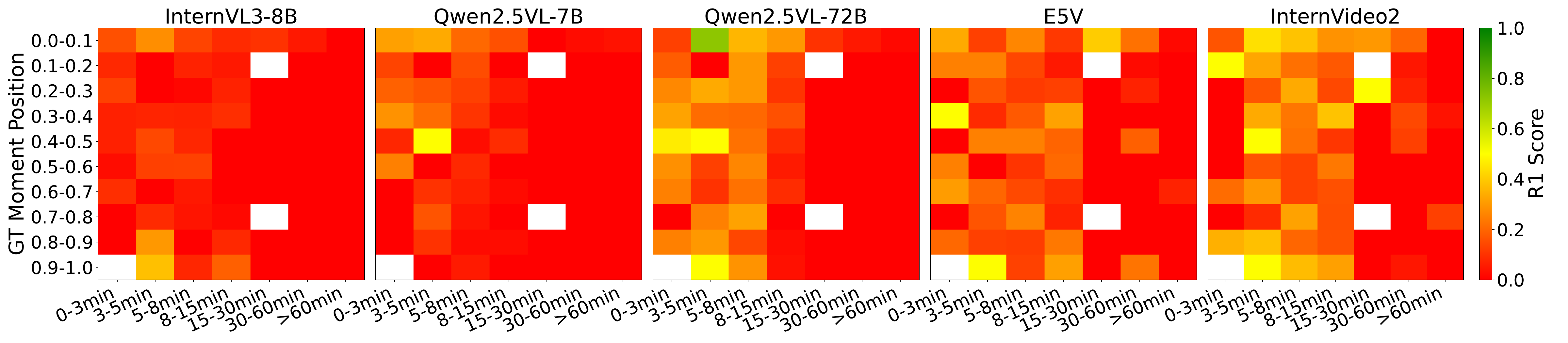}
\caption{Accuracy of retrieval methods $w.r.t$ ground-truth interval position and video length. Generation methods better predict start/end intervals, while retrieval methods are less affected by interval position. All models decline as video duration grows.}
  \label{fig:heatmap}
\end{figure*}

\textbf{1. MomentSeeker remains highly challenging.}
Even advanced MLLMs such as GPT-4o and Gemini 2.5 Pro achieve relatively low scores on our benchmark, highlighting the difficulty of fine-grained temporal grounding. The full-set performance of GPT-4o remains largely consistent with prior results, while Gemini 2.5 Pro achieves the best reported numbers (29.6 R@1 and 31.4 mAP@5) yet still struggles on queries requiring precise temporal localization (e.g., counting the number of goals). This underscores the inherent complexity of MomentSeeker and its potential to reveal subtle temporal reasoning weaknesses in modern MLLMs.

\textbf{2. Existing methods struggle with multi-modal queries.}
Figure~\ref{fig:modality} shows that most models perform worse on IMR (Image-conditioned Moment Retrieval) and VMR (Video-conditioned Moment Retrieval) than on TMR (Text-only Moment Retrieval). This gap is especially large for generation-based methods, indicating difficulty in integrating and reasoning across modalities. While current models can handle simpler cross-modal tasks (e.g., cross-modal retrieval) or less complex multi-modal understanding tasks (e.g., basic video understanding) reasonably well, they often fail to capture the deeper relationships required for complex multi-modal reasoning. This limitation hinders their robustness and generalizability in real-world scenarios.

\textbf{3. Generation methods are position sensitive, while retrieval methods are not.}
Figure~\ref{fig:heatmap} illustrates model sensitivity to video duration and ground-truth interval positions. InternVL3-8B often predicts intervals near the start or end, and Qwen2.5VL-7B shows a strong bias toward the beginning. Qwen2.5VL-72B mitigates this issue with a more balanced prediction distribution. Retrieval models, treating all chunks equally regardless of position, are largely position-insensitive and consistently generate near-uniform predictions. Moreover, all models' performance generally declines as video duration increases: retrieval models face a larger candidate pool, complicating ranking, while generation models suffer greater information loss due to more aggressive downsampling, reducing prediction accuracy.

\textbf{4. Generation-based models are significantly constrained by context length.}
As shown in Figure~\ref{fig:heatmap}, Qwen2.5VL-72B achieves noticeably higher accuracy on short videos compared to strong retrievers such as E5V and InternVideo2. This is primarily because Qwen2.5VL-72B supports up to 768 input frames, which corresponds to approximately 6.4 minutes of video when sampled at 2 fps. For videos under 8 minutes, the information loss due to uniform frame downsampling is minimal. Moreover, generation-based methods can ingest the full context, which is especially crucial for many global-level tasks. As a result, they outperform chunking-based retrieval methods. We further confirm that longer temporal context consistently improves performance. For instance, Eagle2.5-8B (256-frame input) outperforms InternVL3-8B (96-frame input) by 2.77 overall, reinforcing the value of extended context for temporal reasoning. These results suggest that with sufficiently long context support, generation-based models have a much higher potential ceiling.

\textbf{5. Generation methods lag behind retrieval methods, but scale helps.}
As shown in Table~\ref{tab:main_exp}, most generation-based methods lag behind retrieval-based methods in moment retrieval performance. For example, the lightweight retriever MM-Ret (148M) achieves an overall score of 12.4 R@1 and 17.7 mAP@5, significantly outperforming the much larger generation model InternVL3-8B, which only achieves 5.9 R@1 and 6.1 mAP@5. However, as model size increases (e.g., InternVL3-38B), generation-based methods begin to approach the performance of retrieval-based ones. This indicates that while current MLLMs lack strong temporal reasoning, increasing model size helps close the gap.

\textbf{6. MomentSeeker performance predicts downstream LVU task success.}
We use different retrievers (Qwen2.5VL, MM-Ret, InternVideo2) in a RAG pipeline to select top-$k$ key moments, then feed them to the same MLLM (InternVL3-38B) for LVU tasks. Figure~\ref{fig:retriever} shows a positive correlation between MomentSeeker retriever performance and downstream RAG-based LVU results, indicating our benchmark effectively predicts LVU task capability.

In summary, LVMR remains a challenging problem given the limitations in overall performance. Retrieval methods generally outperform generation-based methods, but both fall short in handling fine-grained temporal reasoning and complex multi-modal queries. While scaling up generation models and extending temporal context both help, even the latest video MLLMs still struggle on MomentSeeker, underscoring the benchmark's difficulty and diagnostic value. Notably, moment retrieval quality strongly correlates with LVU task performance, highlighting MomentSeeker’s value as an intermediate benchmark.

\section{Conclusion}
We present MomentSeeker, a new benchmark designed to address the challenge of accurately locating key moments in LVU for MLLMs. MomentSeeker covers a wide range of real-world tasks and supports various query types: text-only, image-conditioned, and video-conditioned. Extensive experiments on both generation-based and retrieval-based methods reveal significant challenges in accuracy and efficiency, despite improvements brought by recent MLLMs and task-specific fine-tuning. We hope MomentSeeker will serve as a valuable resource to advance the development of more accurate and efficient LVU systems. We discuss the limitations and future work in Appendix~\ref{sec:limitation}.

\newpage

\section*{Acknowledgments}
This work was supported by Beijing Natural Science Foundation No. L233008 and National Natural Science Foundation of China No. 62272467. The work was partially done at the Engineering Research Center of Next-Generation Intelligent Search and Recommendation, MOE.

{
\small
\bibliographystyle{plain}
\bibliography{egbib}
}

\newpage

\section{Appendix}
\label{sec:appendix}

\subsection{Evaluation Setting}\label{app:evaluation}

We conduct all the experiments on 8×A800 GPUs, each with 80 GB of memory.
\subsubsection{Retrieval-based Methods}~\label{app:retrieval_setting}
Following previous works~\cite{jiang2025vlm2vec, jiang2024e5}, we incorporated instructions (e.g., ``Represent the given question in one word.'' and ``Represent the video in one word.'') to guide the model in generating informative embeddings for queries and candidates.  All baselines are reproduced using their official implementations with default settings. We evaluate the video models (LanguageBind and InternVideo2) using an input of uniformly sampled 8 frames, while COVR follows its default setting of 15 frames. For image models (E5V, VLM2Vec, UniIR and MagicLens), we use the temporally middle frame as the video input. Instruction usage remains consistent with each baseline’s original configuration. For image-based baselines, we use the middle frame of the video as input. CoVR~\cite{ventura2024covr} follows its original configuration with 15 frames at a resolution of 384, while LanguageBind, InternVideo2~\cite{wang2025internvideo2}, and MR-Embedder each use 8 frames at a resolution of 224.  For image-based models (MM-Ret, MagicLens, UniIR, VLM2VEC, E5V), we use the middle frame of videos as input.

\subsubsection{Generation-based Methods}~\label{app:generation_setting}
All evaluation code for the generation-based methods is sourced from the official GitHub repository. The number of input frames for each model is based on the optimal frame count referenced in the original papers for long video tasks. The prompts we used are as follows:

For GPT-4o, we randomly sampled 180 test samples as the testing set. We extracted video frames at 0.5 fps, and for videos exceeding 128 frames, we uniformly sampled 128 frames. The videos were resized so that the maximum length or width did not exceed 384.

For Lita, we use the default setting: videos are extended by 100 frames without resizing. The prompt to Lita is as follows. First, we provide the timestamps of the sampled frames to Lita, allowing it to infer the temporal positions of each frame.

\begin{tcolorbox}[promptstyle, title={\texttt{Time Instruction}}]
Video 1 lasts for {:.2f} seconds, and \{\} frames are uniformly sampled from it.  
These frames correspond to the following timestamps: \{\}. Please answer the following questions based on this video.
\end{tcolorbox}

Next, we provide task-specific prompts based on TMR, IMR, and VMR tasks.

For the TMR task, the following prompt is used. Depending on whether the input is a query or a caption, the prompt wording varies slightly.

\begin{tcolorbox}[promptstyle, title={\texttt{TMR TASK PROMPT}}]
Identify the visual event described by the following query/caption in the video, and determine its start and end times.  
Format: 'The event happens in the start time - end time'.  
Example: The event 'person turns on a light' happens in the 24.3 - 30.4 seconds.  
Now, here is the textual query: \{\}. Please return its start and end times.
\end{tcolorbox}

For TimeChat in the default setting, videos are uniformly sampled to 96 frames. The prompts for different input types in the TMR task are as follows:

\begin{tcolorbox}[promptstyle, title={\texttt{TimeChat Query-Type TMR Task}}]
You are given a video from the QVHighlights dataset. Identify the visual event described by the given sentence, and determine its start and end times.  
Format: 'The event happens in the start time - end time'.  
Example: The event 'person turns on a light' happens in the 24.3 - 30.4 seconds.  
Here is the sentence: \{\}. Please return its start and end times.
\end{tcolorbox}

\begin{tcolorbox}[promptstyle, title={\texttt{TimeChat Caption-Type TMR Task}}]
Detect and report the start and end timestamps of the video segment that semantically matches the given sentence.  
Format: 'The event happens in the start time - end time'.  
Example: The event 'person turns on a light' happens in the 24.3 - 30.4 seconds.  
Here is the sentence: \{\}. Please return its start and end times.
\end{tcolorbox}

For LLaVA-Video, InternVL3, and Qwen2.5VL, we also use the default setting. Videos are sampled to 96 frames for LLaVA-Video and InternVL3, and 768 frames for Qwen2.5VL. LLaVA-Video and Qwen2.5VL share the same time instruction format as Lita. InternVL3 includes timestamps directly before each \texttt{<image>} token. All models use the same prompt template as below:

\begin{tcolorbox}[promptstyle, title={\texttt{LLaVA-Video / InternVL3 / Qwen2.5VL: TMR Task}}]
Identify the most relevant time interval(s) in the video that match the given query or caption.  
Format: [[start\_1, end\_1], ..., [start\_n, end\_n]], where $1 \leq n \leq 5$.

Examples:  
Single interval: [[0.2, 7.8]]  
Multiple intervals: [[0, 10.3], [65.4, 67.3]]

Now, here is the textual query: \{\}

\textbf{IMPORTANT:}  
1. Return \textbf{only} the list of relevant intervals in Video 1.  
2. Do not return more than 5 intervals.
\end{tcolorbox}

\begin{tcolorbox}[promptstyle, title={\texttt{LLaVA-Video / InternVL3 / Qwen2.5VL: IMR Task}}]
Identify one or more time intervals in the video that match the given query paired with an image.  
Format: [[start\_1, end\_1], ..., [start\_n, end\_n]], where $1 \leq n \leq 5$.

Examples:  
Single interval: [[0.2, 7.8]]  
Multiple intervals: [[0, 10.3], [65.4, 67.3]]

Here is the textual query: \{\} and the accompanying image: Image 1.

\textbf{IMPORTANT:}  
1. Return \textbf{only} the list of relevant intervals in Video 1.  
2. Do not return more than 5 intervals.
\end{tcolorbox}

\begin{tcolorbox}[promptstyle, title={\texttt{LLaVA-Video / InternVL3 / Qwen2.5VL: VMR Task}}]
Identify one or more time intervals in the video that match the given query paired with a reference video.  
Format: [[start\_1, end\_1], ..., [start\_n, end\_n]], where $1 \leq n \leq 5$.

Examples:  
Single interval: [[0.2, 7.8]]  
Multiple intervals: [[0, 10.3], [65.4, 67.3]]

Here is the textual query: \{\} and the accompanying video: Video 2.

\textbf{IMPORTANT:}  
1. Return \textbf{only} the list of relevant intervals in Video 1.  
2. Do not return more than 5 intervals.
\end{tcolorbox}

\subsection*{Annotation Guideline}\label{app:annotation}

\begin{Verbatim}
--- Task 1: Video Search (MR) ---

Objective: Generate a question answerable by a specific video segment and annotate the 
corresponding one or more answering segments.

Scope: Both the question and answer must originate from the same video.

Steps:

1. Watch the entire video to understand its content (e.g., actions, events, objects).
2. Formulate a question:
   - Ensure the question is specific and requires one or several continuous segments as 
   the answer.
   - Example questions:
     * "Where was the weighing scale?"
     * "What did I put in the trashcan?"
3. Annotate the answer segment:
   - Mark the start and end timestamps (format: [[MM:SS--MM:SS], [MM:SS--MM:SS]...]).
   - Ensure the segments fully answer the question without truncation.
   - Cover all segments that answer the question—no omissions allowed.
4. Validation: Replay the annotated segment to confirm alignment with the question.

--- Task 2: Image-Conditioned Video Search (IMR) ---

Objective: Generate a question based on a static image (from the same or a different 
video) and annotate the answer segments in the target video.

Steps:
1. Select an image:
   - Same video: Choose a key frame (e.g., action initiation, critical object appearance).
   - Different video: Use a frame from another video with relevant content (e.g., 
   similar objects, scenes).
2. Formulate a question:
   - The question must directly relate to the selected image, and the answer must 
   exist in the target video.
   - Example questions:
     * Same video (image shows a road with street scenery): "What’s the color of 
     the dog that once appeared on this road?"
     * Different video (image shows the same character in different clothing): 
     "What did the man in this picture give to the person next to him?"
3. Annotate the answer segment:
   - Mark the timestamps in the target video that answer the question.
   - Ensure logical consistency between the image, question, and segment.
4. Validation: Verify that the image, question, and annotated segment are coherent.

--- Quality Requirements ---

1. Consistency: Ensure IMR questions are tightly linked to the image, and 
answers are precise.
2. Cross-video logic: If using an image from another video, the question must 
relate to the target video’s content (avoid mixing contexts).
3. Timestamp accuracy: Annotated segments must have less than 1-second error tolerance.

Report ambiguities or edge cases to the project lead for resolution.
\end{Verbatim}

\subsection{Ablation of Different Video Frame Numbers}~\label{app:frame_result}
In the main experiments, we followed the optimal frame settings officially recommended for each model. Additionally, Table~\ref{tab:main_exp} presents the performance of different models under various frame input conditions.We observe that varying the number of input frames within a moderate range from 64 to 128 has minimal impact on performance. For example, InternVL3-38B achieves stable overall performance with R@1 values of 15.9, 15.8, and 15.6 from 64 to 128 frames, suggesting that denser sampling in this range does not significantly improve results. But increasing the number of frames from 96 to 768 for Qwen2.5VL-7B results in a marked performance gain, from 4.5 to 8.1 in R@1. This suggests that models benefit more from denser temporal coverage, likely due to their limited ability to infer information from sparsely sampled inputs.However, such improvements result in a considerable computational cost for processing 768 frames significantly increases memory and latency.

\begin{table*}[t!]
\centering
\caption{Main results across different meta-tasks. \#Frames indicates the number of input frames for generation-based methods and per-clip frames for retrieval-based methods.  $\dag$ denotes tested on a random subset due to high cost. }
\label{tab:ablation_meta_task}
\resizebox{\textwidth}{!}{%
\begin{tabular}{
  l|cc|
  cc@{\hskip 2pt}>{\columncolor{white}}c@{\hskip 2pt}  
  cc@{\hskip 2pt}>{\columncolor{white}}c@{\hskip 2pt}
  cc@{\hskip 2pt}>{\columncolor{white}}c@{\hskip 2pt}
  cc}
\toprule
\multirow{2}{*}{\textbf{Method}} & 
\multirow{2}{*}{\textbf{\#Size}} & 
\multirow{2}{*}{\textbf{\#Frames}} & 
\multicolumn{2}{c}{\textbf{Global-level}} & &
\multicolumn{2}{c}{\textbf{Event-level}} & &
\multicolumn{2}{c}{\textbf{Object-level}} & &
\multicolumn{2}{c}{\textbf{Overall}} \\

\cmidrule{4-5} \cmidrule{7-8} \cmidrule{10-11} \cmidrule{13-14}  
 & & & R@1 & mAP@5 & 
\multicolumn{1}{c}{} & 
R@1 & mAP@5 & 
\multicolumn{1}{c}{} & 
R@1 & mAP@5 & 
\multicolumn{1}{c}{} & 
R@1 & mAP@5 \\

\midrule
\multicolumn{14}{c}{\textit{\textbf{Generation-based Methods}}} \\
\hline
Qwen2.5VL & 7B & 96 & \cellgreen{1.8} & \cellgreen{1.5} & & \cellgreen{7.0} & \cellgreen{7.0} & & \cellgreen{2.4} & \cellgreen{2.1} & & \cellgreen{4.5} & \cellgreen{4.3} \\
InternVL3 & 8B & 96 & \cellgreen{3.9} & \cellgreen{3.5} & & \cellgreen{7.8} & \cellgreen{8.5} & & \cellgreen{4.1} & \cellgreen{4.1} & & \cellgreen{5.9} & \cellgreen{6.1} \\
InternVL3 & 38B & 64 & \cellgreen{11.1} & \cellgreen{10.5} & & \cellgreen{20.9} & \cellgreen{21.3} & & \cellgreen{11.3} & \cellgreen{11.5} & & \cellgreen{15.9} & \cellgreen{16.0} \\
InternVL3 & 38B & 96 & \cellgreen{11.1} & \cellgreen{10.5} & & \cellgreen{20.8} & \cellgreen{21.2} & & \cellgreen{11.3} & \cellgreen{11.5} & & \cellgreen{15.8} & \cellgreen{16.0} \\
InternVL3 & 38B & 128 & \cellgreen{11.1} & \cellgreen{10.5} & & \cellgreen{20.1} & \cellgreen{20.5} & & \cellgreen{11.3} & \cellgreen{11.4} & & \cellgreen{15.6} & \cellgreen{15.7} \\
Qwen2.5VL & 7B & 256 & \cellgreen{2.3} & \cellgreen{2.2} & & \cellgreen{6.8} & \cellgreen{7.2} & & \cellgreen{2.0} & \cellgreen{2.1} & & \cellgreen{4.4} & \cellgreen{4.6} \\
Qwen2.5VL & 7B & 768 & \cellgreen{4.6} & \cellgreen{3.8} & & \cellgreen{12.0} & \cellgreen{12.2} & & \cellgreen{4.3} & \cellgreen{4.2} & & \cellgreen{8.1} & \cellgreen{8.0} \\
Qwen2.5VL & 7B & 64 & \cellgreen{2.6} & \cellgreen{2.1} & & \cellgreen{6.8} & \cellgreen{7.0} & & \cellgreen{2.2} & \cellgreen{2.2} & & \cellgreen{4.5} & \cellgreen{4.5} \\
Qwen2.5VL & 7B & 128 & \cellgreen{2.8} & \cellgreen{2.4} & & \cellgreen{7.6} & \cellgreen{7.8} & & \cellgreen{2.0} & \cellgreen{1.9} & & \cellgreen{4.9} & \cellgreen{4.9} \\

\bottomrule
\end{tabular}
}
\end{table*}

\subsection{Ablation of Different IoU Thresholds}\label{app:iou}
We provide experimental results under different IoU settings (0.1,0.2,0.4,0.5) in Table~\ref{tab:iou0.1}, Table~\ref{tab:iou0.2}, Table~\ref{tab:iou0.4}, and Table~\ref{tab:iou0.5}. It can be observed that stricter IoU thresholds (e.g., 0.5) lead to a drop in the accuracy of all models. However, the experimental conclusions drawn in Section~\ref{sec:experiments} still hold under different IoU settings.

\begin{table*}[t!]
\centering
\caption{Results of IoU=0.1.}
\label{tab:iou0.1}
\resizebox{\textwidth}{!}{%
\begin{tabular}{
  l|cc|
  cc@{\hskip 2pt}>{\columncolor{white}}c@{\hskip 2pt}  
  cc@{\hskip 2pt}>{\columncolor{white}}c@{\hskip 2pt}
  cc@{\hskip 2pt}>{\columncolor{white}}c@{\hskip 2pt}
  cc}
\toprule
\multirow{2}{*}{\textbf{Method}} & 
\multirow{2}{*}{\textbf{\#Size}} & 
\multirow{2}{*}{\textbf{\#Frames}} & 
\multicolumn{2}{c}{\textbf{Global-level}} & &
\multicolumn{2}{c}{\textbf{Event-level}} & &
\multicolumn{2}{c}{\textbf{Object-level}} & &
\multicolumn{2}{c}{\textbf{Overall}} \\

\cmidrule{4-5} \cmidrule{7-8} \cmidrule{10-11} \cmidrule{13-14}  
 & & & R@1 & mAP@5 & 
\multicolumn{1}{c}{} & 
R@1 & mAP@5 & 
\multicolumn{1}{c}{} & 
R@1 & mAP@5 & 
\multicolumn{1}{c}{} & 
R@1 & mAP@5 \\

\midrule
\multicolumn{14}{c}{\textit{\textbf{Generation-based Methods}}} \\
\hline
InternVL3 & 8B & 96 & \cellgreen{6.4} & \cellgreen{5.8} & & \cellgreen{13.8} & \cellgreen{15.0} & & \cellgreen{7.1} & \cellgreen{7.1} & & \cellgreen{10.3} & \cellgreen{10.7} \\
InternVL3 & 38B & 96 & \cellgreen{23.9} & \cellgreen{22.0} & & \cellgreen{31.2} & \cellgreen{31.8} & & \cellgreen{20.2} & \cellgreen{19.7} & & \cellgreen{26.3} & \cellgreen{26.1} \\
Qwen2.5VL & 7B & 768 & \cellgreen{9.8} & \cellgreen{8.8} & & \cellgreen{22.5} & \cellgreen{22.9} & & \cellgreen{9.1} & \cellgreen{9.0} & & \cellgreen{15.7} & \cellgreen{15.7} \\
TimeChat & 7B & 96 & \cellgreen{5.3} & \cellgreen{5.3} & & \cellgreen{16.0} & \cellgreen{16.0} & & \cellgreen{6.6} & \cellgreen{6.6} & & \cellgreen{13.1} & \cellgreen{13.1} \\
Lita & 13B & 100 & \cellgreen{11.8} & \cellgreen{11.8} & & \cellgreen{18.3} & \cellgreen{18.3} & & \cellgreen{9.3} & \cellgreen{9.3} & & \cellgreen{15.8} & \cellgreen{15.8} \\
Qwen2.5VL & 72B & 768 & \cellgreen{22.6} & \cellgreen{21.2} & & \cellgreen{30.7} & \cellgreen{30.7} & & \cellgreen{19.9} & \cellgreen{19.0} & & \cellgreen{25.7} & \cellgreen{25.1} \\
LLaVA-Video & 72B & 96 & \cellgreen{8.5} & \cellgreen{8.4} & & \cellgreen{16.4} & \cellgreen{18.0} & & \cellgreen{8.3} & \cellgreen{9.8} & & \cellgreen{12.3} & \cellgreen{13.5} \\
\hline
\multicolumn{14}{c}{\textit{\textbf{Retrieval-based Methods}}} \\
\hline
E5V & 8.4B & 1 & \cellgreen{21.6} & \cellgreen{28.6} & & \cellgreen{23.5} & \cellgreen{33.0} & & \cellgreen{26.7} & \cellgreen{33.2} & & \cellgreen{24.1} & \cellgreen{32.0} \\
COVR & 588M & 15 & \cellgreen{14.4} & \cellgreen{21.4} & & \cellgreen{23.0} & \cellgreen{31.7} & & \cellgreen{21.3} & \cellgreen{28.0} & & \cellgreen{20.4} & \cellgreen{28.0} \\
InternVideo2& 1B & 8 & \cellgreen{29.1} & \cellgreen{38.1} & & \cellgreen{38.8} & \cellgreen{48.7} & & \cellgreen{27.4} & \cellgreen{35.5} & & \cellgreen{32.7} & \cellgreen{41.7} \\
UniIR & 428M & 1 & \cellgreen{24.5} & \cellgreen{29.6} & & \cellgreen{21.6} & \cellgreen{30.9} & & \cellgreen{20.0} & \cellgreen{27.2} & & \cellgreen{21.7} & \cellgreen{29.3} \\
MM-Ret & 148M & 1 & \cellgreen{19.3} & \cellgreen{25.8} & & \cellgreen{23.3} & \cellgreen{31.1} & & \cellgreen{17.5} & \cellgreen{25.5} & & \cellgreen{20.4} & \cellgreen{27.9} \\
LanguageBind & 428M & 8 & \cellgreen{30.4} & \cellgreen{39.3} & & \cellgreen{38.1} & \cellgreen{48.0} & & \cellgreen{27.2} & \cellgreen{35.0} & & \cellgreen{32.6} & \cellgreen{41.5} \\
\bottomrule
\end{tabular}
}
\end{table*}

\begin{table*}[t!t!]
\centering
\caption{Results of IoU=0.2.}
\label{tab:iou0.2}
\resizebox{\textwidth}{!}{%
\begin{tabular}{
  l|cc|
  cc@{\hskip 2pt}>{\columncolor{white}}c@{\hskip 2pt}  
  cc@{\hskip 2pt}>{\columncolor{white}}c@{\hskip 2pt}
  cc@{\hskip 2pt}>{\columncolor{white}}c@{\hskip 2pt}
  cc}
\toprule
\multirow{2}{*}{\textbf{Method}} & 
\multirow{2}{*}{\textbf{\#Size}} & 
\multirow{2}{*}{\textbf{\#Frames}} & 
\multicolumn{2}{c}{\textbf{Global-level}} & &
\multicolumn{2}{c}{\textbf{Event-level}} & &
\multicolumn{2}{c}{\textbf{Object-level}} & &
\multicolumn{2}{c}{\textbf{Overall}} \\

\cmidrule{4-5} \cmidrule{7-8} \cmidrule{10-11} \cmidrule{13-14}  
 & & & R@1 & mAP@5 & 
\multicolumn{1}{c}{} & 
R@1 & mAP@5 & 
\multicolumn{1}{c}{} & 
R@1 & mAP@5 & 
\multicolumn{1}{c}{} & 
R@1 & mAP@5 \\

\midrule
\multicolumn{14}{c}{\textit{\textbf{Generation-based Methods}}} \\
\hline
IntennVL3 & 8B & 96 & \cellgreen{4.4} & \cellgreen{4.0} & & \cellgreen{10.9} & \cellgreen{12.0} & & \cellgreen{5.8} & \cellgreen{5.7} & & \cellgreen{8.0} & \cellgreen{8.4} \\
IntennVL3 & 38B & 96 & \cellgreen{18.0} & \cellgreen{16.7} & & \cellgreen{25.6} & \cellgreen{26.2} & & \cellgreen{15.8} & \cellgreen{15.6} & & \cellgreen{21.0} & \cellgreen{20.9} \\
Qwen2.5VL & 7B & 768 & \cellgreen{7.2} & \cellgreen{6.2} & & \cellgreen{16.7} & \cellgreen{17.2} & & \cellgreen{6.7} & \cellgreen{6.6} & & \cellgreen{11.7} & \cellgreen{11.7} \\
TimeChat & 7B & 96 & \cellgreen{3.9} & \cellgreen{3.9} & & \cellgreen{10.5} & \cellgreen{10.5} & & \cellgreen{5.3} & \cellgreen{5.3} & & \cellgreen{8.8} & \cellgreen{8.8} \\
Lita & 13B & 100 & \cellgreen{7.9} & \cellgreen{7.9} & & \cellgreen{10.6} & \cellgreen{10.6} & & \cellgreen{4.9} & \cellgreen{4.9} & & \cellgreen{9.1} & \cellgreen{9.1} \\
Qwen2.5VL & 72B & 768 & \cellgreen{17.5} & \cellgreen{16.4} & & \cellgreen{26.5} & \cellgreen{26.4} & & \cellgreen{15.8} & \cellgreen{15.2} & & \cellgreen{21.3} & \cellgreen{20.9} \\
LLaVA-Video & 72B & 96 & \cellgreen{4.9} & \cellgreen{4.7} & & \cellgreen{12.0} & \cellgreen{13.5} & & \cellgreen{6.1} & \cellgreen{7.1} & & \cellgreen{8.7} & \cellgreen{9.7} \\
\hline
\multicolumn{14}{c}{\textit{\textbf{Retrieval-based Methods}}} \\
\hline
E5V & 8.4B & 1 & \cellgreen{19.6} & \cellgreen{25.9} & & \cellgreen{20.4} & \cellgreen{28.8} & & \cellgreen{20.7} & \cellgreen{26.9} & & \cellgreen{20.3} & \cellgreen{27.5} \\
COVR & 588M & 15 & \cellgreen{13.1} & \cellgreen{19.5} & & \cellgreen{19.7} & \cellgreen{27.9} & & \cellgreen{17.4} & \cellgreen{23.5} & & \cellgreen{17.4} & \cellgreen{24.4} \\
UniIR & 428M & 1 & \cellgreen{21.6} & \cellgreen{26.3} & & \cellgreen{19.2} & \cellgreen{27.3} & & \cellgreen{14.4} & \cellgreen{21.3} & & \cellgreen{18.2} & \cellgreen{25.1} \\
MM-Ret & 148M & 1 & \cellgreen{17.5} & \cellgreen{23.0} & & \cellgreen{20.1} & \cellgreen{27.3} & & \cellgreen{14.0} & \cellgreen{21.3} & & \cellgreen{17.4} & \cellgreen{24.3} \\
LanguageBind & 428M & 8 & \cellgreen{25.0} & \cellgreen{33.6} & & \cellgreen{32.3} & \cellgreen{42.6} & & \cellgreen{21.1} & \cellgreen{28.2} & & \cellgreen{26.8} & \cellgreen{35.6} \\
\bottomrule
\end{tabular}
}
\end{table*}

\begin{table*}[t!]
\centering
\caption{Results of IoU=0.4.}
\label{tab:iou0.4}
\resizebox{\textwidth}{!}{%
\begin{tabular}{
  l|cc|
  cc@{\hskip 2pt}>{\columncolor{white}}c@{\hskip 2pt}  
  cc@{\hskip 2pt}>{\columncolor{white}}c@{\hskip 2pt}
  cc@{\hskip 2pt}>{\columncolor{white}}c@{\hskip 2pt}
  cc}
\toprule
\multirow{2}{*}{\textbf{Method}} & 
\multirow{2}{*}{\textbf{\#Size}} & 
\multirow{2}{*}{\textbf{\#Frames}} & 
\multicolumn{2}{c}{\textbf{Global-level}} & &
\multicolumn{2}{c}{\textbf{Event-level}} & &
\multicolumn{2}{c}{\textbf{Object-level}} & &
\multicolumn{2}{c}{\textbf{Overall}} \\

\cmidrule{4-5} \cmidrule{7-8} \cmidrule{10-11} \cmidrule{13-14}  
 & & & R@1 & mAP@5 & 
\multicolumn{1}{c}{} & 
R@1 & mAP@5 & 
\multicolumn{1}{c}{} & 
R@1 & mAP@5 & 
\multicolumn{1}{c}{} & 
R@1 & mAP@5 \\

\midrule
\multicolumn{14}{c}{\textit{\textbf{Generation-based Methods}}} \\
\hline
InternVL3 & 8B & 96 & \cellgreen{1.3} & \cellgreen{1.2} & & \cellgreen{5.8} & \cellgreen{6.3} & & \cellgreen{3.2} & \cellgreen{3.2} & & \cellgreen{4.1} & \cellgreen{4.3} \\
InternVL3 & 38B & 96 & \cellgreen{7.7} & \cellgreen{7.5} & & \cellgreen{16.1} & \cellgreen{16.3} & & \cellgreen{8.2} & \cellgreen{8.1} & & \cellgreen{11.9} & \cellgreen{12.0} \\
Qwen2.5VL & 7B & 768 & \cellgreen{3.6} & \cellgreen{2.9} & & \cellgreen{8.9} & \cellgreen{9.0} & & \cellgreen{2.8} & \cellgreen{2.9} & & \cellgreen{5.9} & \cellgreen{5.8} \\
TimeChat & 7B & 96 & \cellgreen{1.3} & \cellgreen{1.3} & & \cellgreen{4.9} & \cellgreen{4.9} & & \cellgreen{3.5} & \cellgreen{3.5} & & \cellgreen{4.3} & \cellgreen{4.3} \\
Lita & 13B & 100 & \cellgreen{1.3} & \cellgreen{1.3} & & \cellgreen{4.0} & \cellgreen{4.0} & & \cellgreen{1.3} & \cellgreen{1.3} & & \cellgreen{3.2} & \cellgreen{3.2} \\
Qwen2.5VL & 72B & 768 & \cellgreen{10.0} & \cellgreen{9.4} & & \cellgreen{15.2} & \cellgreen{15.2} & & \cellgreen{9.1} & \cellgreen{8.6} & & \cellgreen{12.3} & \cellgreen{12.0} \\
LLaVA-Video & 72B & 96 & \cellgreen{2.8} & \cellgreen{2.6} & & \cellgreen{5.0} & \cellgreen{6.1} & & \cellgreen{4.1} & \cellgreen{4.6} & & \cellgreen{4.3} & \cellgreen{4.9} \\
\hline
\multicolumn{14}{c}{\textit{\textbf{Retrieval-based Methods}}} \\
\hline
E5V & 8.4B & 1 & \cellgreen{6.2} & \cellgreen{10.5} & & \cellgreen{10.9} & \cellgreen{15.9} & & \cellgreen{10.6} & \cellgreen{15.1} & & \cellgreen{9.7} & \cellgreen{14.3} \\
COVR & 588M & 15 & \cellgreen{6.2} & \cellgreen{9.2} & & \cellgreen{10.6} & \cellgreen{15.8} & & \cellgreen{10.5} & \cellgreen{14.4} & & \cellgreen{9.5} & \cellgreen{13.7} \\
InternVideo2& 1B & 8 & \cellgreen{10.6} & \cellgreen{15.5} & & \cellgreen{18.3} & \cellgreen{23.9} & & \cellgreen{14.4} & \cellgreen{18.6} & & \cellgreen{15.1} & \cellgreen{20.1} \\
UniIR & 428M & 1 & \cellgreen{8.8} & \cellgreen{11.0} & & \cellgreen{8.3} & \cellgreen{13.6} & & \cellgreen{5.6} & \cellgreen{9.9} & & \cellgreen{7.5} & \cellgreen{11.8} \\
MM-Ret & 148M & 1 & \cellgreen{6.4} & \cellgreen{9.4} & & \cellgreen{10.6} & \cellgreen{15.0} & & \cellgreen{7.6} & \cellgreen{12.2} & & \cellgreen{8.6} & \cellgreen{12.7} \\
LanguageBind & 428M & 8 & \cellgreen{10.3} & \cellgreen{15.3} & & \cellgreen{16.7} & \cellgreen{23.4} & & \cellgreen{11.8} & \cellgreen{16.4} & & \cellgreen{13.5} & \cellgreen{19.1} \\
\bottomrule
\end{tabular}
}
\end{table*}

\begin{table*}[t!]
\centering
\caption{Results of IoU=0.5.}
\label{tab:iou0.5}
\resizebox{\textwidth}{!}{%
\begin{tabular}{
  l|cc|
  cc@{\hskip 2pt}>{\columncolor{white}}c@{\hskip 2pt}  
  cc@{\hskip 2pt}>{\columncolor{white}}c@{\hskip 2pt}
  cc@{\hskip 2pt}>{\columncolor{white}}c@{\hskip 2pt}
  cc}
\toprule
\multirow{2}{*}{\textbf{Method}} & 
\multirow{2}{*}{\textbf{\#Size}} & 
\multirow{2}{*}{\textbf{\#Frames}} & 
\multicolumn{2}{c}{\textbf{Global-level}} & &
\multicolumn{2}{c}{\textbf{Event-level}} & &
\multicolumn{2}{c}{\textbf{Object-level}} & &
\multicolumn{2}{c}{\textbf{Overall}} \\

\cmidrule{4-5} \cmidrule{7-8} \cmidrule{10-11} \cmidrule{13-14}  
 & & & R@1 & mAP@5 & 
\multicolumn{1}{c}{} & 
R@1 & mAP@5 & 
\multicolumn{1}{c}{} & 
R@1 & mAP@5 & 
\multicolumn{1}{c}{} & 
R@1 & mAP@5 \\

\midrule
\multicolumn{14}{c}{\textit{\textbf{Generation-based Methods}}} \\
\hline
InternVL3 & 8B & 96 & \cellgreen{1.0} & \cellgreen{0.9} & & \cellgreen{4.0} & \cellgreen{4.2} & & \cellgreen{2.8} & \cellgreen{2.8} & & \cellgreen{3.0} & \cellgreen{3.1} \\
InternVL3 & 38B & 96 & \cellgreen{4.6} & \cellgreen{4.4} & & \cellgreen{12.6} & \cellgreen{12.8} & & \cellgreen{6.3} & \cellgreen{6.1} & & \cellgreen{9.0} & \cellgreen{9.0} \\
Qwen2.5VL & 7B & 768 & \cellgreen{3.1} & \cellgreen{2.4} & & \cellgreen{6.7} & \cellgreen{6.6} & & \cellgreen{1.9} & \cellgreen{1.9} & & \cellgreen{4.4} & \cellgreen{4.3} \\
TimeChat & 7B & 96 & \cellgreen{1.3} & \cellgreen{1.3} & & \cellgreen{2.9} & \cellgreen{2.9} & & \cellgreen{1.8} & \cellgreen{1.8} & & \cellgreen{2.5} & \cellgreen{2.5} \\
Lita & 13B & 100 & \cellgreen{0.0} & \cellgreen{0.0} & & \cellgreen{2.1} & \cellgreen{2.1} & & \cellgreen{1.3} & \cellgreen{1.3} & & \cellgreen{1.8} & \cellgreen{1.8} \\
Qwen2.5VL & 72B & 768 & \cellgreen{6.9} & \cellgreen{6.2} & & \cellgreen{11.8} & \cellgreen{11.8} & & \cellgreen{5.9} & \cellgreen{5.5} & & \cellgreen{9.0} & \cellgreen{8.7} \\
LLaVA-Video & 72B & 96 & \cellgreen{1.3} & \cellgreen{1.1} & & \cellgreen{2.6} & \cellgreen{3.5} & & \cellgreen{3.2} & \cellgreen{3.6} & & \cellgreen{2.5} & \cellgreen{3.0} \\
\hline
\multicolumn{14}{c}{\textit{\textbf{Retrieval-based Methods}}} \\
\hline
E5V & 8.4B & 1 & \cellgreen{2.8} & \cellgreen{5.3} & & \cellgreen{7.8} & \cellgreen{11.7} & & \cellgreen{6.9} & \cellgreen{10.3} & & \cellgreen{6.3} & \cellgreen{9.7} \\
COVR & 588M & 15 & \cellgreen{3.9} & \cellgreen{6.1} & & \cellgreen{7.7} & \cellgreen{11.6} & & \cellgreen{7.3} & \cellgreen{10.1} & & \cellgreen{6.6} & \cellgreen{9.8} \\
InternVideo2& 1B & 8 & \cellgreen{5.7} & \cellgreen{8.5} & & \cellgreen{14.2} & \cellgreen{18.2} & & \cellgreen{10.8} & \cellgreen{13.0} & & \cellgreen{11.0} & \cellgreen{14.1} \\
UniIR & 428M & 1 & \cellgreen{4.6} & \cellgreen{5.8} & & \cellgreen{6.6} & \cellgreen{10.3} & & \cellgreen{3.2} & \cellgreen{6.4} & & \cellgreen{5.0} & \cellgreen{7.9} \\
MM-Ret & 148M & 1 & \cellgreen{3.1} & \cellgreen{4.8} & & \cellgreen{6.8} & \cellgreen{10.3} & & \cellgreen{5.0} & \cellgreen{7.8} & & \cellgreen{5.3} & \cellgreen{8.1} \\
LanguageBind & 428M & 8 & \cellgreen{4.4} & \cellgreen{8.4} & & \cellgreen{12.3} & \cellgreen{17.0} & & \cellgreen{7.8} & \cellgreen{10.5} & & \cellgreen{8.9} & \cellgreen{12.8} \\
\bottomrule
\end{tabular}
}
\end{table*}

\subsection{Visualization Results}\label{app:visual}

\begin{figure*}[t!]
  \centering
  \includegraphics[width=0.7\linewidth]{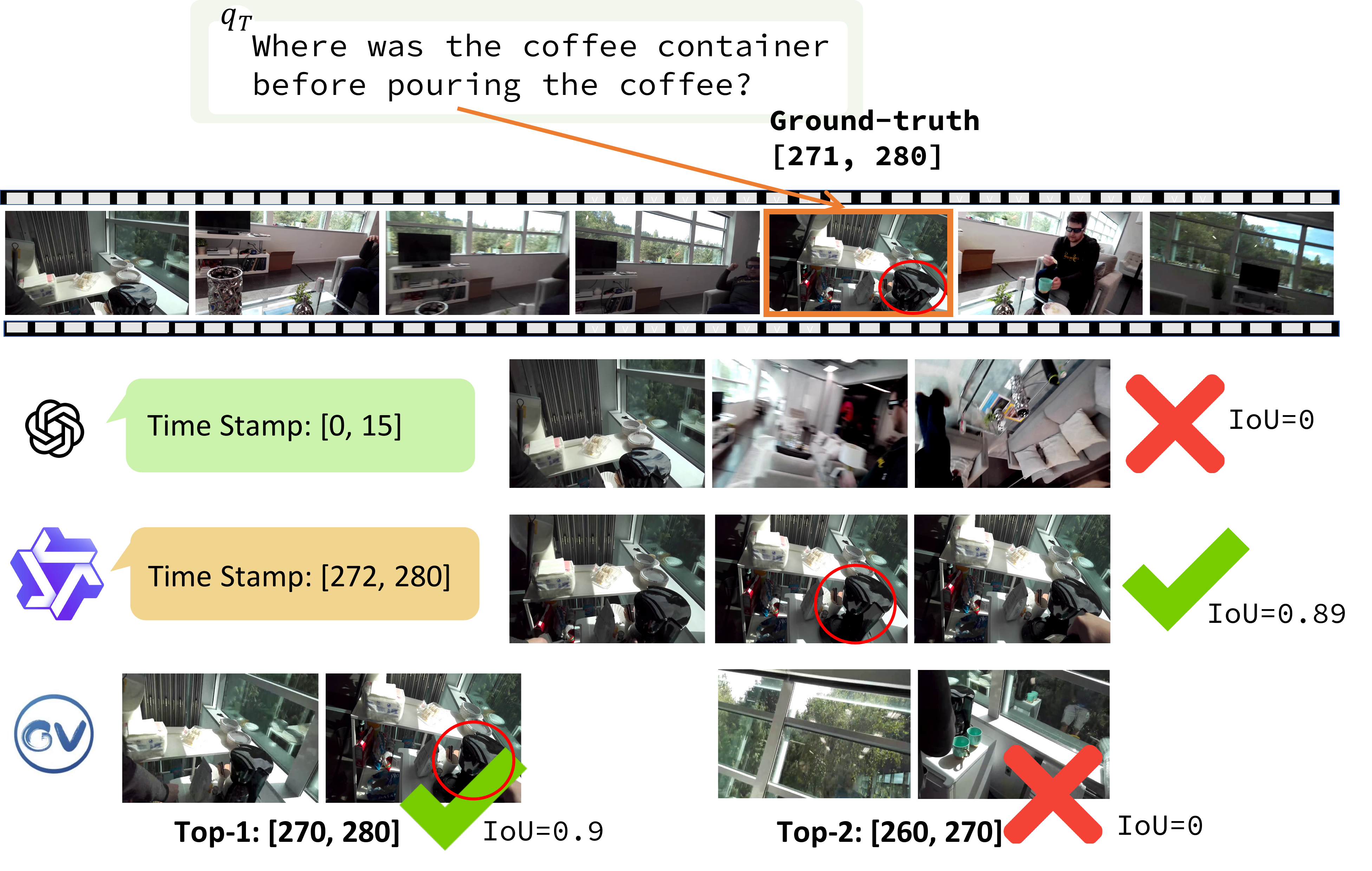}

\caption{Case Study of generation-based methods (GPT-4o, Qwen2.5VL-72B), and retrieval-based model InternVideo2.}
  \label{fig:app_demo1}
\end{figure*}

\begin{figure*}[t!]
  \centering
  \includegraphics[width=0.7\linewidth]{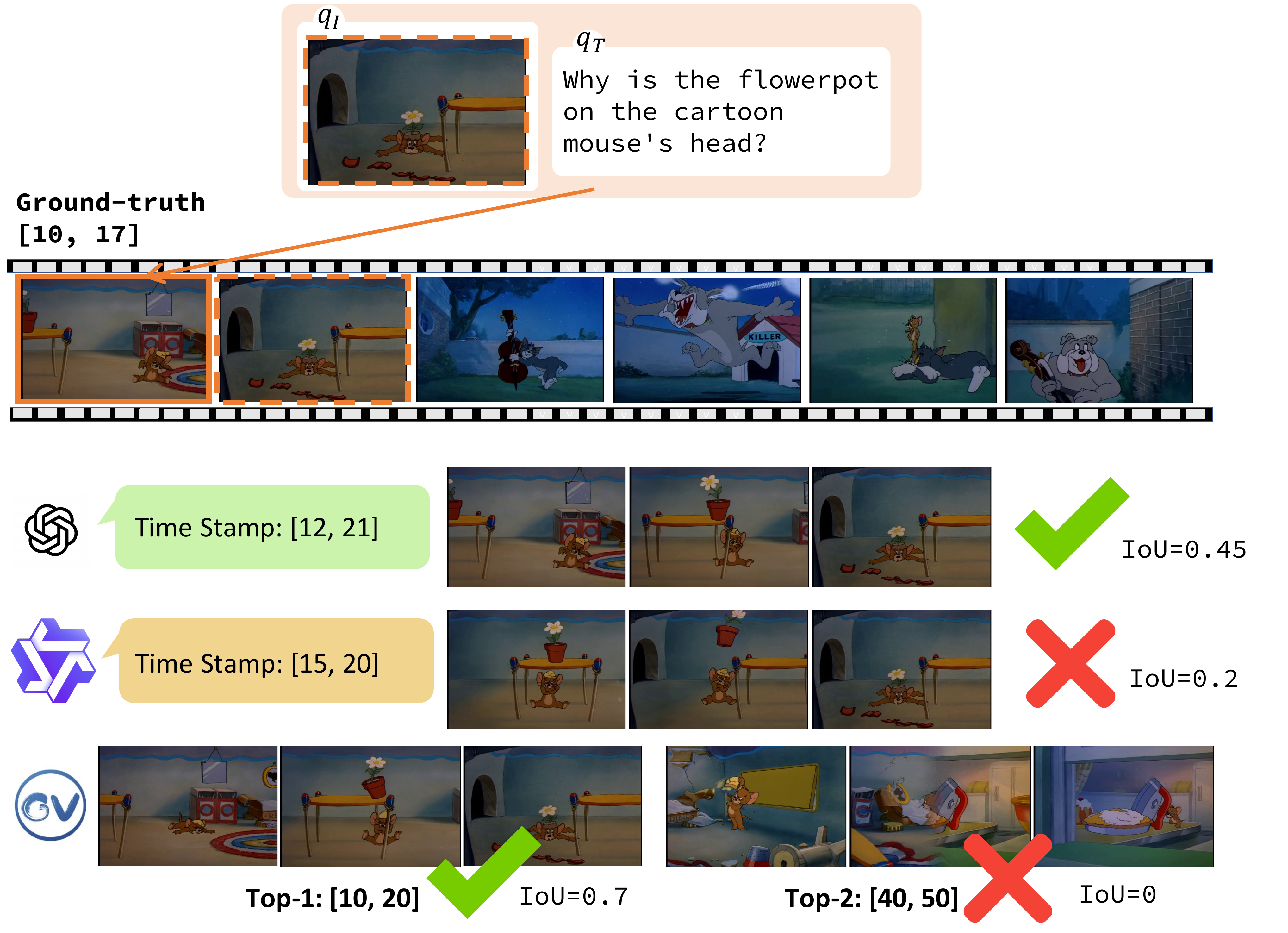}

\caption{Case Study of generation-based methods (GPT-4o, Qwen2.5VL-72B), and retrieval-based model InternVideo2.}
  \label{fig:app_demo2}
\end{figure*}

\begin{figure*}[t!]
  \centering
  \includegraphics[width=0.7\linewidth]{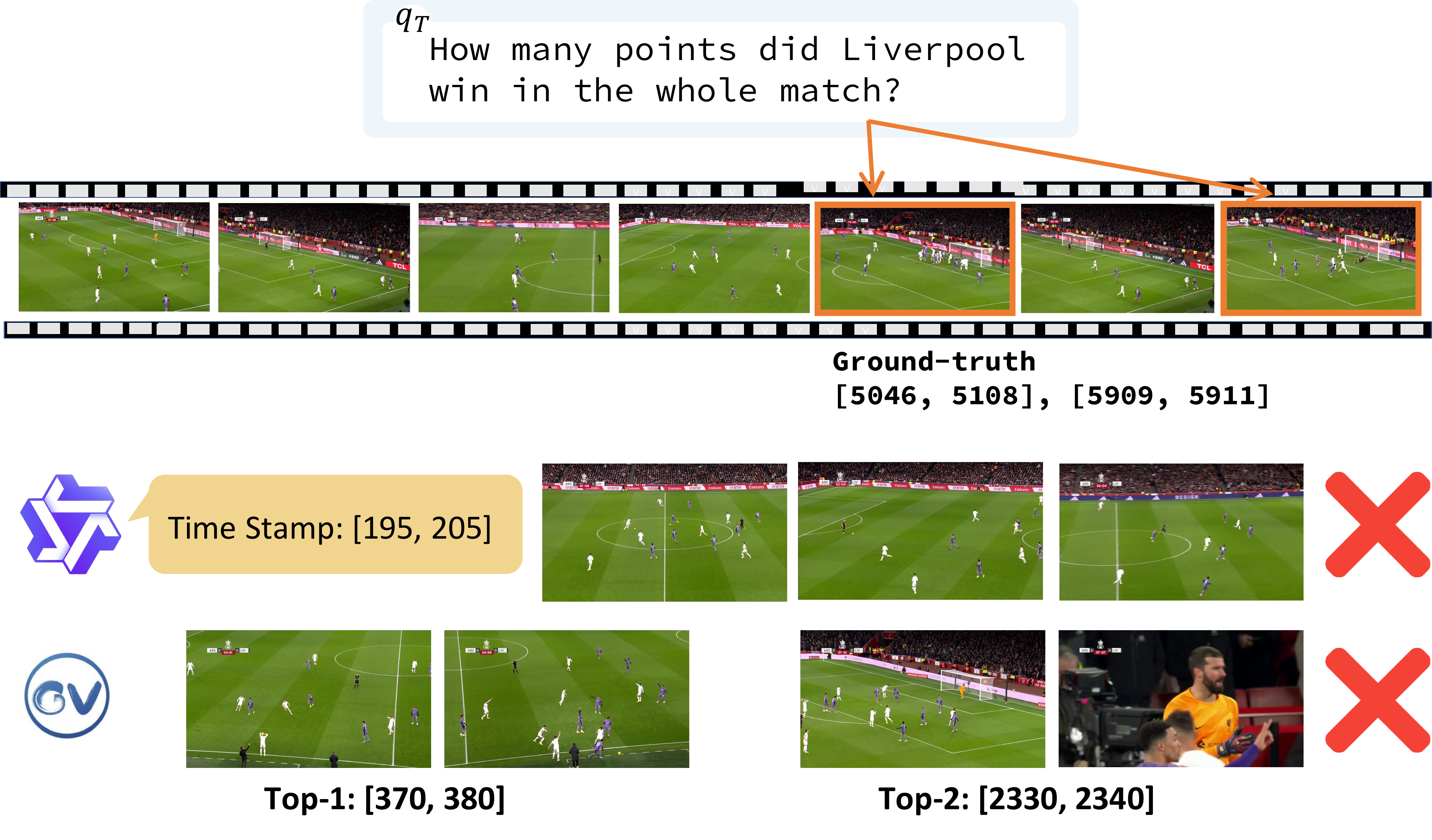}

\caption{Case Study of generation-based methods (GPT-4o, Qwen2.5VL-7B), and retrieval-based model InternVideo2.}
 \label{fig:app_demo3}
\end{figure*}

We provide case study of generation-based methods (GPT-4o, Qwen2.5VL-72B), and retrieval-based model InternVideo2 in Figure~\ref{fig:app_demo1}, Figure~\ref{fig:app_demo2} and Figure~\ref{fig:app_demo3}. 

\section*{Broader Impacts}\label{app:ethics}
All the videos in our benchmark are carefully extracted from publicly available datasets ~\cite{mangalam2023egoschema,zhou2025mlvu,wang2024lvbench,sultani2019anomaly}, which have undergone rigorous screening by the original teams to remove harmful content. Despite our best efforts, we acknowledge that these videos may not be entirely comprehensive or free from omissions. Furthermore, we strongly discourage the use of MR-Embedder models for encoding or retrieving sensitive content.

\section{Limitations \& Future Works}\label{sec:limitation}

\textbf{Limitations.} This paper focuses on moment retrieval in long videos and presents a comprehensive benchmark evaluation. We conduct extensive experiments from multiple perspectives, including task types, query modalities, video lengths, and moment distributions. While we tested a proprietary MLLM (GPT-4o) in our study, large-scale evaluation on long videos with closed-source models remains prohibitively expensive. For instance, processing 384 frames at 512 resolution with GPT-4o costs approximately \$2 per query, amounting to around \$3,600 for our dataset of 1,800 queries. Due to this high cost, we did not explore other proprietary MLLMs in this work and instead focused primarily on open-source models. In addition, although we provide extensive empirical evaluations of existing methods, we do not delve into model-specific architectural choices, such as the design of temporal embeddings or structural adaptations to enhance temporal reasoning in MLLMs.

\textbf{Future work suggestions.} Two promising directions emerge for future long-video understanding models. First, current MLLMs show performance drops on fine-grained temporal localization~\cite{fu2024videomme,zhou2025mlvu}; thus, efforts to enhance their temporal awareness could be beneficial. Second, integrating lightweight retrievers into RAG frameworks may improve efficiency by focusing on key video segments. By advancing both directions together, we may improve the accuracy of existing LVU models on downstream tasks by enhancing their temporal awareness.

\end{document}